\documentclass[11pt]{article}

\usepackage[preprint]{acl}

\usepackage{times}
\usepackage{latexsym}

\usepackage[T1]{fontenc}

\usepackage[utf8]{inputenc}

\usepackage{microtype}

\usepackage{inconsolata}

\usepackage{graphicx}

\usepackage{xspace}
\usepackage{pifont}
\newcommand{\ourmethod}{{\fontfamily{lmtt}\selectfont \textbf{NeuReasoner}}\xspace}
\newcommand{\fmon}{{\fontfamily{lmtt}\selectfont \textbf{Mixture of Neurons}}\xspace}
\newcommand{\mon}{{\fontfamily{lmtt}\selectfont \textbf{MoN}}\xspace}
\newcommand{\lk}{{\fontfamily{lmtt}\selectfont \textbf{(}}\xspace}
\newcommand{\rk}{{\fontfamily{lmtt}\selectfont \textbf{)}}\xspace}
\definecolor{ccr}{RGB}{72, 192, 170}
\usepackage{twemojis}
\usepackage[table,xcdraw,dvipsnames]{xcolor}
\usepackage{booktabs}
\usepackage{makecell}
\usepackage{bbding}
\usepackage{multirow}
\usepackage{array}
\newcommand{\darkred}[1]{\textcolor{RedOrange}{\ensuremath{\uparrow}\,#1}}
\newcommand{\darkblue}[1]{\textcolor{BlueGreen}{\ensuremath{\downarrow}\,#1}}
\definecolor{MorandiHeader}{RGB}{219, 226, 236} 
\definecolor{AltRowColor}{RGB}{245,244,242}
\definecolor{DeltaRowColor}{RGB}{236, 244, 221}
\newcommand{\gcell}[1]{\cellcolor{AltRowColor}#1}
\newcommand{\dcell}[1]{\cellcolor{DeltaRowColor}#1}
\definecolor{BaselineTag}{RGB}{238,220,210} 
\definecolor{RegenTag}{RGB}{214,230,242}

\definecolor{intra}{HTML}{E07B54}
\definecolor{inter}{HTML}{E1C855}
\definecolor{instance}{HTML}{51B1B7}

\usepackage[most]{tcolorbox}
\definecolor{fancyTeal}{HTML}{008080}
\definecolor{softBack}{HTML}{F0F8FF}
\newtcolorbox{cbx}[1][]{
  enhanced,
  breakable,
  title={#1},
  colframe=fancyTeal,
  colback=softBack,
  colbacktitle=fancyTeal,
  fonttitle=\bfseries\sffamily,
  boxrule=0.5mm,
  arc=3mm,
  attach boxed title to top left={
    yshift=-2mm,
    xshift=4mm
  },
  boxed title style={
    boxrule=0mm,
    arc=1.5mm,
    shadow={2mm}{-2mm}{0mm}{black!15}
  },
  drop fuzzy shadow
}
\usepackage{fontawesome6}
\usepackage{listings}

\lstdefinestyle{PromptStyle}{
    basicstyle=\fontfamily{pcr}\selectfont\small,
    breaklines=true,        
    columns=fullflexible,   
    keepspaces=true,        
    showstringspaces=false, 
    extendedchars=true,     
    frame=none,             
    aboveskip=0pt,          
    belowskip=0pt,
    escapechar=             
}
\newtcolorbox{PromptFrame}[1]{
    enhanced,
    breakable,              
    title={#1},             
    colframe=fancyTeal,
    colback=softBack,
    colbacktitle=fancyTeal,
    fonttitle=\bfseries\sffamily,
    boxrule=0.5mm,
    arc=3mm,
    attach boxed title to top left={
        yshift=-2mm,
        xshift=4mm
    },
    boxed title style={
        boxrule=0mm,
        arc=1.5mm,
        shadow={2mm}{-2mm}{0mm}{black!15}
    },
    drop fuzzy shadow
}

\usepackage[ruled,vlined,linesnumbered]{algorithm2e}
\newcommand{\AlgComment}[1]{{\fontfamily{pcr}\selectfont\bfseries\color{blue}/* #1 */}}
\newcommand{\EqRef}[1]{\hfill \texttt{\textbf{$\triangleright$ Eq. \ref{#1}}}}
\newcommand{\SecRef}[1]{\hfill \texttt{\textbf{$\triangleright$ Sec. \ref{#1}}}}

\usepackage{wrapfig}
\usepackage{enumitem}
\usepackage{amsmath}
\usepackage{amssymb}
\usepackage{mathtools}
\usepackage{amsthm}

\title{\ourmethod: Towards Explainable, Controllable, and Unified Reasoning via Mixture-of-Neurons}

\author{
{\bfseries Haonan Dong$^{1*}$ \quad Kehan Jiang$^{2*}$ \quad Haoran Ye$^{1}$} \\
{\bfseries Wenhao Zhu$^{1}$ \quad Zhaolu Kang$^{2}$ \quad Guojie Song$^{1\dagger}$} \\
$^{1}$State Key Laboratory of General Artificial Intelligence, \\
School of Intelligence Science and Technology, Peking University \\
$^{2}$School of Software and Microelectronics, Peking University \\
$^{*}$Equal contribution \qquad $^{\dagger}$Corresponding author \\
{\faEnvelope\ hndong25@stu.pku.edu.cn, gjsong@pku.edu.cn}
}

\begin{document}
\maketitle
\begin{abstract}
Large Reasoning Models (LRMs) have recently achieved remarkable success in complex reasoning tasks. However, closer scrutiny reveals persistent failure modes compromising \textit{\underline{performance}} and \textit{\underline{cost}}: \textcolor{intra}{\textbf{I) \textit{Intra-step level}}}, marked by calculation or derivation errors; \textcolor{inter}{\textbf{II) \textit{Inter-step level}}}, involving oscillation and stagnation; and \textcolor{instance}{\textbf{III) \textit{Instance level}}}, causing maladaptive over-thinking. Existing endeavors target isolated levels without \textit{unification}, while their black-box nature and reliance on RL hinder \textit{explainability} and \textit{controllability}. To bridge these gaps, we conduct an in-depth white-box analysis, identifying key neurons (\fmon, \mon) and their fluctuation patterns associated with distinct failures. Building upon these insights, we propose \ourmethod, an \textit{\underline{explainable}}, \textit{\underline{controllable}}, and \textit{\underline{unified}} reasoning framework driven by \mon. Technically, \ourmethod integrates lightweight MLPs for failure detection with a special token-triggered self-correction mechanism learned via SFT. During inference, special tokens are inserted upon failure detection to actuate controllable remedial behaviors. Extensive evaluations across six benchmarks, six backbone models (8B$\sim$70B) against nine competitive baselines, demonstrate that \ourmethod achieves performance gains of up to $27.0\%$ while reducing token consumption by $19.6\%\sim63.3\%$.
\end{abstract}

\section{Introduction}
Reasoning capability stands as a fundamental cornerstone of human intelligence and serves as an essential pathway toward AGI~\citep{human-reasoning}. Recently, facilitated by Chain-of-Thought (CoT) techniques~\citep{cot}, Large Language Models (LLMs) demonstrate remarkable potential in complex tasks, such as mathematics~\citep{cot-math}, coding~\citep{cot-code}, and science~\citep{cot-sci}, by employing step-by-step reasoning processes. With the advent of Large Reasoning Models (LRMs) such as OpenAI-o1~\citep{o1}, DeepSeek-R1~\citep{r1}, and Gemini-2.5 pro~\citep{gemini}, Reinforcement Learning (RL)-driven training paradigms further expand the boundaries of reasoning~\citep{rl-1,rl-2,dapo}. This paradigm shift elicit the spontaneous emergence of human-like cognitive patterns within models, including self-verification, reflection, and multi-path exploration~\citep{r1}.

\begin{figure}[!t]
  \centering
  \includegraphics[width=\linewidth]{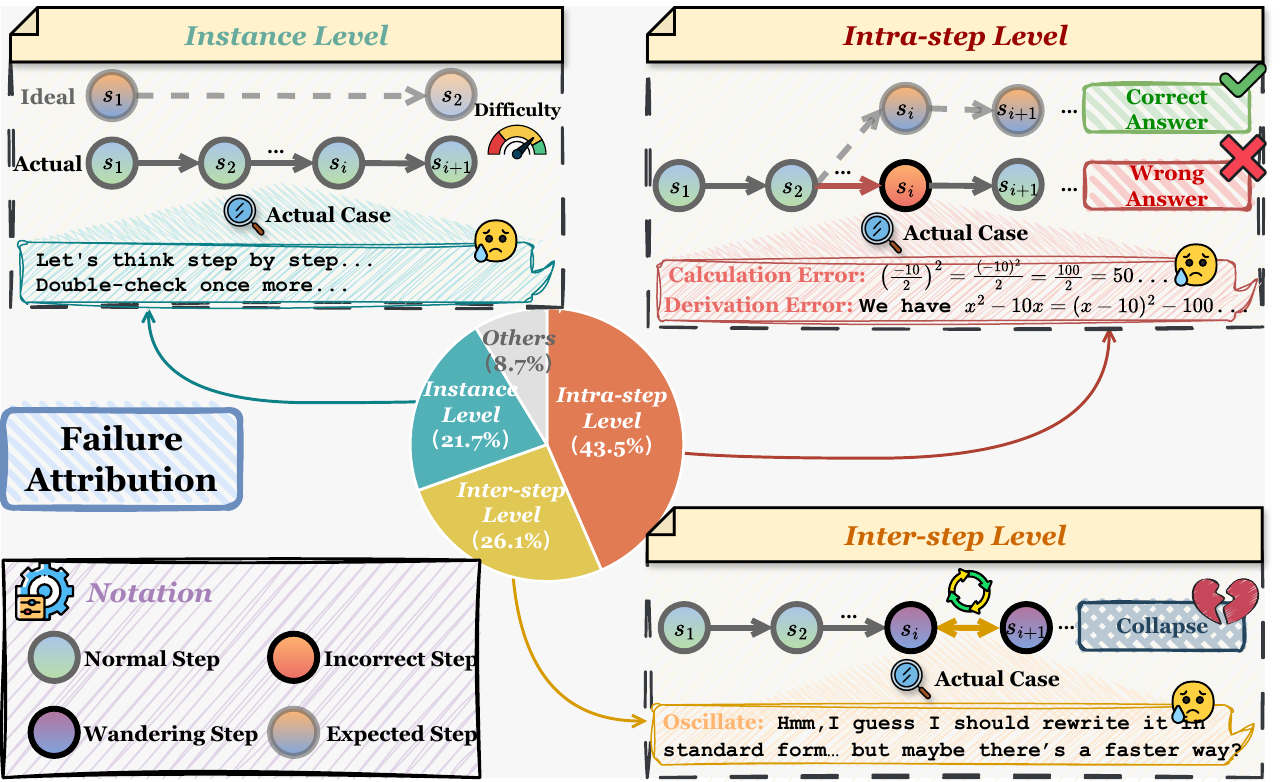}
  \vspace{-1.8em}
  \caption{Distribution and illustration of failure modes across different levels using DeepSeek-R1-Distill-Qwen-7B on MATH~\citep{math}.}
   \label{fig:intro}
   \vspace{-1.5em}
\end{figure}

\paragraph{Failure Attribution.} Despite the remarkable success of LRMs, a close look from the dual perspective of \textit{\underline{performance}} and \textit{\underline{cost}} reveals distinct failure modes concealed within the reasoning process, which we systematically identify (Figure \ref{fig:intro}). Specifically, \textcolor{intra}{\textbf{I) \textit{Intra-step level}}}: flaws in causal deduction or deviation into incorrect branches during critical steps compromise the final answer~\citep{intra}; \textcolor{inter}{\textbf{II) \textit{Inter-step level}}}: models risk entrapment in reasoning stagnation, where they oscillate between similar trajectories without substantive progress, thereby consuming excessive tokens before eventual collapse~\citep{inter}; and \textcolor{instance}{\textbf{III) \textit{Instance level}}}: the failure to calibrate query difficulty induces over-thinking, resulting in a substantial waste of token costs~\citep{instance}.

\paragraph{Research Dilemma.} Prior studies attempt to address these failure modes \textit{individually}, where \textcolor{intra}{\textbf{I)}} Process Reward Models (PRMs) such as Math-Shepherd~\citep{mathshepherd}, AutoPRM~\citep{autoprm}, and OpenPRM~\citep{openprm} are employed for fine-grained step-level supervision; \textcolor{inter}{\textbf{II)}} structured frameworks like ToT~\citep{tot}, GoT~\citep{got}, and PGTS~\citep{pgts} extend the breadth and depth of exploration; and \textcolor{instance}{\textbf{III)}} approaches including O1-Pruner~\citep{o1-pruner}, AdaptThink~\citep{adaptthink}, and ARM~\citep{arm} adaptively adjust reasoning length. However, significant research gaps persist across three critical axes: \ding{182} \textbf{Explainability}, due to the black-box nature of models, in-depth analysis regarding the internal mechanisms underlying failure modes remains absent; \ding{183} \textbf{Controllability}, RL-based paradigms encourage the spontaneous emergence of behaviors, rendering the reasoning process unpredictable and difficult to control; and \ding{184} \textbf{Unification}, existing methods lack a unified solution targeting these failure modes collectively.

\paragraph{Present Framework.} To bridge these gaps, inspired by cognitive science findings that distinct brain regions govern specific functions~\citep{brain-cog}, we hypothesize that three key neuron clusters within LRMs contribute significantly to the three levels of failure modes. Through in-depth analysis of neuron activation dynamics during reasoning, we identify these clusters, termed \fmon \lk\mon\rk, where each cluster functions as an \textit{expert}. Furthermore, we scrutinize the fluctuation patterns of each expert under its corresponding failure mode. Building upon this systematic white-box analysis paradigm, we propose \ourmethod, an \textit{\underline{explainable}}, \textit{\underline{controllable}} and \textit{\underline{unified}} reasoning framework driven by \mon. Technically, \ourmethod \texttwemoji{heart suit} trains lightweight MLPs to monitor and predict the fluctuation patterns of each expert; \texttwemoji{heart suit} through dataset reconstruction, leverages SFT to condition the model to utilize special tokens as triggers that elicit specific behavioral patterns, facilitating controllable self-correction; and \texttwemoji{heart suit} leverages parallel MLPs for online monitoring during inference, inserting special tokens upon identifying failure modes to trigger corresponding behaviors, effectively mitigating failures. 
Our contributions are summarized as follows:
\begin{itemize}[leftmargin=*]
\vspace{-0.2em}
\item[\texttwemoji{snowflake}] \textbf{\textit{Insightful Analysis.}} We systematically summarize the failure modes of LRMs, and inspired by cognitive science, conduct an in-depth white-box analysis at a fine-grained neuron level. This leads to the identification of \mon that contribute to failure modes, alongside their fluctuation patterns.
\vspace{-1.5em}
\item[\texttwemoji{snowflake}] \textbf{\textit{Practical Solution.}} We propose \ourmethod, an explainable and controllable unified reasoning framework driven by \mon. It integrates lightweight MLPs for failure detection with a special token-triggered self-correction mechanism learned via SFT. During inference, special tokens are dynamically inserted upon failure detection to actuate controllable remedial behaviors.
\vspace{-0.5em}
\item[\texttwemoji{snowflake}] \textbf{\textit{Experimental Validation.}} Extensive experiments on six complex benchmarks and six backbone models against nine baselines demonstrate that \ourmethod achieves performance gains of up to $27.0\%$ while simultaneously reducing token consumption by $19.6\%\sim63.3\%$. Furthermore, case studies vividly illustrate reasoning details, verifying the explainability and controllability.
\end{itemize}

\begin{figure*}[!t]
  \centering
  \includegraphics[width=\linewidth]{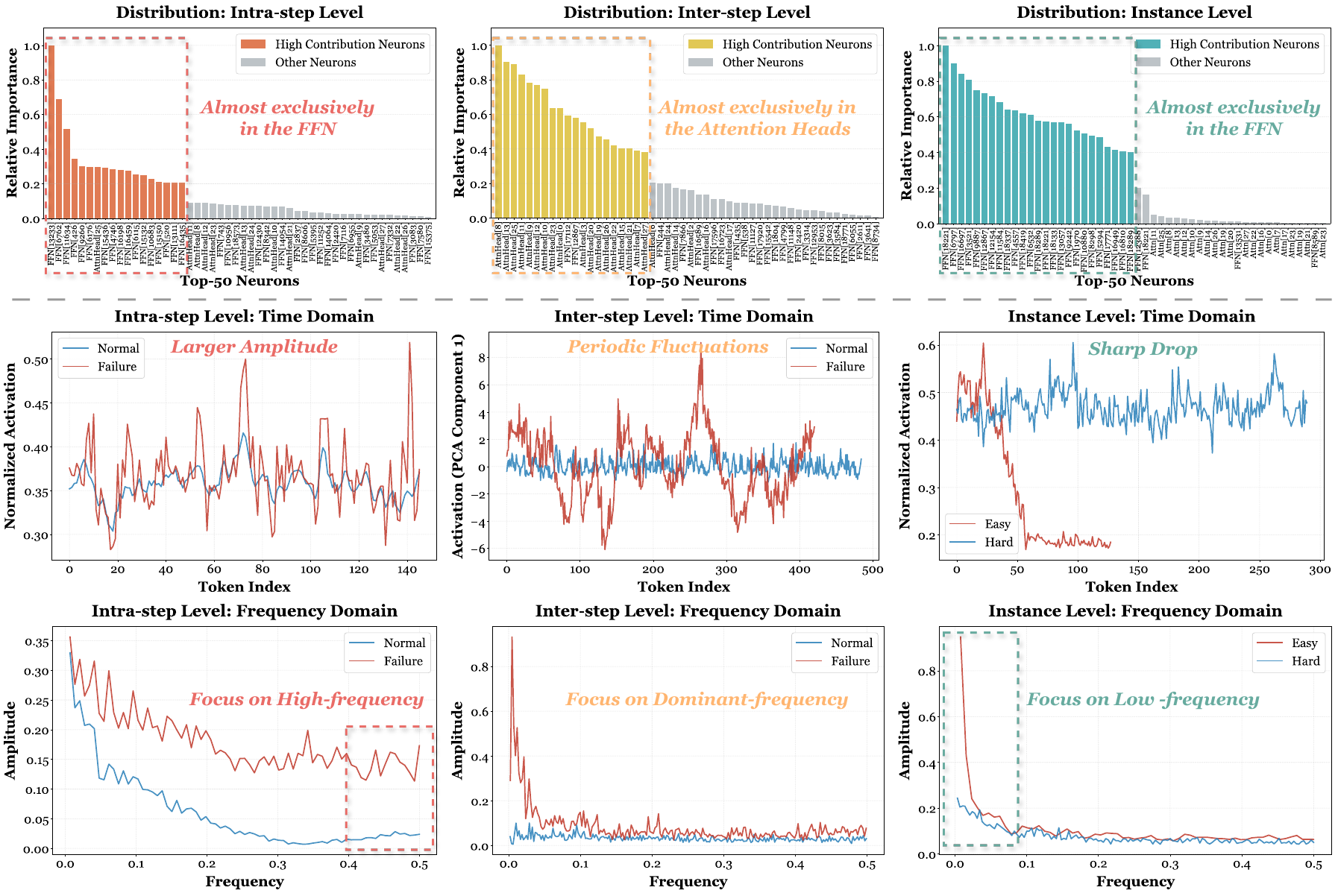}
  \vspace{-2em}
  \caption{(\textbf{\textit{Upper}}) Distribution of key neurons across distinct failure modes using DeepSeek-R1-Distill-Qwen-7B on MATH. (\textbf{\textit{Lower}}) Time- and frequency-domain analysis of \mon for positive and negative sample pairs.}
   \label{fig:mon}
   \vspace{-1.2em}
\end{figure*}

\vspace{-0.7em}
\section{Dive into Neurons}
\label{sec:dive}
Inspired by insights from prior research indicating that intermediate layers within LRMs encode the richest representations~\citep{middle-1,middle-2,middle-5}, we focus our analysis on the FFN and attention heads in the middle layer. Specifically, we perform \textbf{(i) \mon Identification}, leveraging the advanced attribution algorithm of LLMs to identify neuron clusters that contribute most significantly to failure modes of different levels ($\blacktriangleright$ Section \ref{sec:dive-1}); and \textbf{(ii) Fluctuation Analysis}, further uncovering the specific fluctuation patterns of \mon associated with the emergence of failure modes ($\blacktriangleright$ Section \ref{sec:dive-2}).

\subsection{\mon Identification}
\label{sec:dive-1}
To conduct a systematic white-box investigation identifying which specific neuron clusters within the middle layer contribute most significantly to the failure modes of different levels, we employ DePass~\citep{depass}, an advanced attribution algorithm of LLM. Given the target layer $l^* = L/2$, let $\mathcal{C}$ denote the set of all neurons (FFN and attention heads) within layer $l^*$. For a given input sequence, DePass enables the calculation of the independent contribution of each neuron $c \in \mathcal{C}$ toward the token generation at time step $t$. Let $\phi(c, t) \in \mathbb{R}$ denote the attribution score of $c$ regarding the output logit of the target token $y_t$:
\vspace{-0.3em}
\begin{equation}
  \phi(c, t) = \mathbf{w}_{y_t}^\top \mathbf{h}_{dec}^{(l^*)}(t, c), 
\end{equation}
where $\mathbf{w}_{y_t}$ represents the
LM head vector for $y_t$, and $\mathbf{h}_{dec}^{(l^*)}(t, c)$ corresponds to the decomposed hidden state associated with $c$.

\paragraph{\fmon.} We define the set of time steps $T$ as follows: \textcolor{intra}{\textbf{I) \textit{Intra-step level}}}, the sequence of tokens within the specific erroneous step; \textcolor{inter}{\textbf{II) \textit{Inter-step level}}}, the sequence of tokens corresponding to the initial steps of multiple attempts; \textcolor{instance}{\textbf{III) \textit{Instance level}}}, the first $K$ steps at the beginning of the reasoning process. To isolate neurons exerting a persistent and dominant influence across all time steps, we compute the intersection of the most significant neurons $\mathcal{N}^* = \bigcap_{t \in T} \operatorname{TopK}_{c \in \mathcal{C}} \left( \phi(c, t) \right)$, where $\operatorname{TopK}_{c \in \mathcal{C}}(\cdot)$ retrieves the set of the top-$k$ neurons exhibiting the highest attribution scores. The $\mathcal{N}^*$ across all levels collectively constitute what we term \fmon \lk\mon\rk, with each serving as a distinct \textit{expert}, as illustrated in Figure \ref{fig:mon} (\textbf{\textit{Upper}}).
\vspace{-0.4em}
\paragraph{Obs.\ding{182} Distinct neuron experts contribute significantly to failure modes of different levels.} Specifically, experts associated with intra-step and instance-level failures predominantly cluster within the FFN, whereas those corresponding to inter-step failures are concentrated within the attention heads.

\begin{figure*}[!t]
  \centering
  \includegraphics[width=\linewidth]{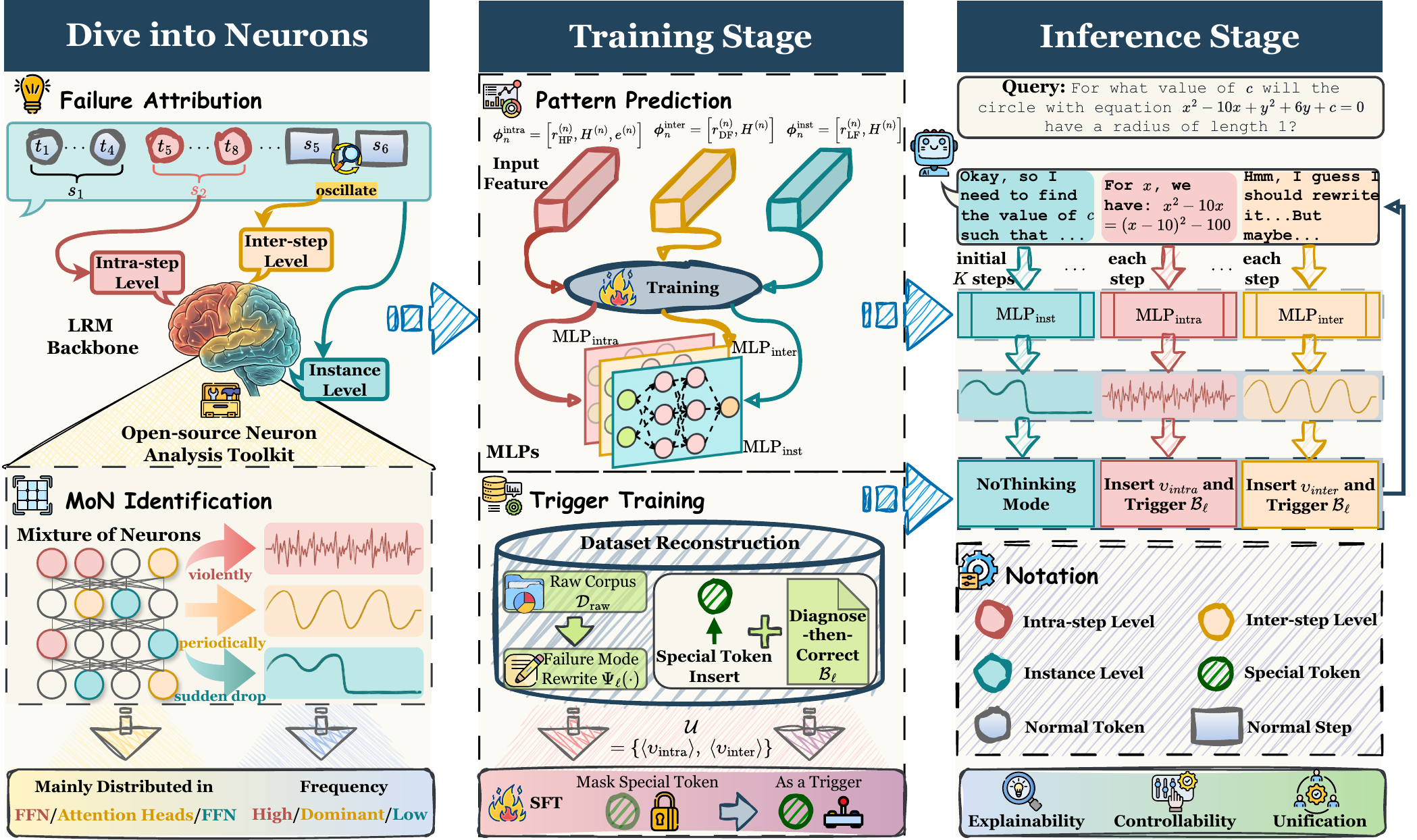}
  \vspace{-1.5em}
  \caption{The overview of our proposed \ourmethod.}
   \label{fig:overview}
   \vspace{-1.2em}
\end{figure*}

\vspace{-0.3em}
\subsection{Fluctuation Analysis}
\label{sec:dive-2}
\vspace{-0.3em}
Following \mon identification, we further aim to analyze the underlying fluctuation patterns associated with the emergence of failure modes. To this end, we construct positive and negative sample pairs. Technically, targeting intra-step and inter-step failures, we pause generation at each step, clone the KV cache, and employ LLM-as-Judge to detect failure modes~\citep{llm-as-judge-2}. Upon detection, we perform repeated sampling from the preceding step to construct the pairs.
For the instance level, 
we construct contrastive instances by selecting two questions with substantially different difficulty. 
Finally, we leverage the fourier transform to analyze fluctuation patterns. Figure \ref{fig:mon} (\textbf{\textit{Lower}}) illustrates the time- and frequency-domain profiles of \mon for sample pairs under different failure modes.
\vspace{-0.4em}
\paragraph{Obs.\ding{183} Corresponding experts within the \mon exhibit distinct fluctuation patterns across failure levels.} Specifically, experts associated with intra-step failures display sharp spikes, whereas those linked to inter-step failures exhibit periodic fluctuations; regarding instance level patterns, experts maintain sustained activation for complex instances but undergo rapid collapse for simpler ones.

\vspace{-0.3em}
\section{Method}
\vspace{-0.3em}
In this section, we elaborate on our proposed framework, \ourmethod. Specifically, \textbf{(i) Pattern Prediction}, building upon the preceding in-depth qualitative analysis of neurons, we train a set of lightweight MLPs to quantify and predict distinct fluctuation patterns ($\blacktriangleright$ Section \ref{sec:method-1}); \textbf{(ii) Trigger Training}, by reconstructing the original dataset according to failure modes, we employ SFT to enable the model to utilize special tokens as triggers to elicit specific behavioral patterns ($\blacktriangleright$ Section \ref{sec:method-2}); and \textbf{(iii) Online Monitoring}, during inference, we deploy MLPs for parallel online monitoring, inserting special tokens upon the emergence of failure modes to induce specific behaviors, thereby achieving self-improvement ($\blacktriangleright$ Section \ref{sec:method-3}).
\vspace{-0.3em}
\subsection{Pattern Prediction}
\label{sec:method-1}
\vspace{-0.3em}
Having identified the fluctuation patterns of failure modes in Section \ref{sec:dive-2}, we aim to train lightweight MLPs to monitor and predict the fluctuations of \mon. Motivated by our observations and prior works~\citep{ts-1,ts-2}, we employ the fourier transform to construct input features.

\vspace{-0.3em}
\paragraph{Preliminary.} For an arbitrary neuron $n$
, we denote its scalar activation values over a token sequence of length $T$ as $a_1, a_2, \ldots, a_T$. Let the activation magnitude be $x_t = |a_t|$, with the corresponding mean given by $\mu=\frac{1}{T}\sum_{t=1}^{T}x_t$. To prevent the mean from dominating the spectrum, we first perform mean removal to obtain the zero-centered sequence $y_t=x_t-\mu$. We apply the Discrete Fourier Transform (DFT)~\citep{dft} to the real-valued sequence $y_t$. Let the number of non-redundant frequency points be $F=\left\lfloor\frac{T}{2}\right\rfloor+1$.
The frequency domain coefficients are defined as $Y(f)=\sum_{t=1}^{T}y_t\exp\!\left(-i\frac{2\pi}{T}(f-1)(t-1)\right)$,
where $f=1,2,\dots,F$ serves as the discrete frequency index. Consequently, the power spectrum is given by $P(f)=|Y(f)|^2$, where $f=1$ corresponds to the Direct Current (DC) component. To construct length-invariant and comparable frequency-domain statistics, we exclude the DC component
to obtain the power distribution:
\vspace{-0.5em}
\begin{equation}
    \widetilde P(f)=\frac{P(f)}{\sum_{j=2}^{F}P(j)+\varepsilon},\, f=2,\dots,F,
\end{equation}
where $\varepsilon>0$ is a constant for numerical stability. Finally, we have the normalized spectral entropy:
\vspace{-0.5em}
\begin{equation}
    H=-\frac{1}{\log(F-1)}\sum_{f=2}^{F}\widetilde P(f)\log\big(\widetilde P(f)\big).
\end{equation}

\paragraph{\textcolor{intra}{I) \textit{Intra-step level}.}} Given that intra-step failure modes are characterized by intense fluctuations, manifesting as stronger high-frequency components and more dispersed spectra, we focus on extracting the high-frequency energy ratio, spectral entropy, and total variation energy. We first define the high-frequency set as the latter half of the spectrum $\mathcal F_{\mathrm{HF}}
=
\left\{\,f\ \middle|\ \left\lfloor\frac{F}{2}\right\rfloor+1 \le f \le F\,\right\}$.
The high-frequency energy ratio is calculated as $r_{\mathrm{HF}}=\sum_{f\in\mathcal F_{\mathrm{HF}}}\widetilde P(f)$. Additionally, we define the total variation energy $e=\log(\sum_{f=2}^{F}P(f)+\varepsilon)$.
Consequently, the input feature is formed as $\boldsymbol{\phi}^{\text{intra}}_n=\big[r_{\mathrm{HF}}^{(n)},\ H^{(n)},\ e^{(n)}\big]\in\mathbb{R}^{3}$.

\paragraph{\textcolor{inter}{II) \textit{Inter-step level.}}} Given that inter-step level failure modes are accompanied by periodic oscillations, signifying the prolonged dominance of energy at a specific non-zero frequency, we focus on extracting the dominant frequency energy ratio and spectral entropy. Specifically, the dominant frequency energy ratio is defined as $r_{\mathrm{dom}}=\max_{2\le f\le F}\widetilde P(f)$. Formally, we have $\boldsymbol{\phi}^{\text{inter}}_n=\big[r_{\mathrm{dom}}^{(n)},\ H^{(n)}\big]\in\mathbb{R}^{2}$.

\paragraph{\textcolor{instance}{III) \textit{Instance level}.}} Regarding instance-level failure modes, compared to hard queries, easy queries exhibit a distinct pattern characterized by initial intense fluctuations followed by a sharp drop and subsequent stabilization. In the frequency domain, this manifests as a concentration of energy in low-frequency components and lower spectral entropy. To quantify this, we first define the low-frequency set as the former half of the spectrum
$\mathcal F_{\mathrm{LF}} = \left\{f\ \middle|\ 2 \le f \le \left\lfloor\frac{F}{2}\right\rfloor\right\}$. The low-frequency energy ratio is calculated as 
$r_{\mathrm{LF}}=\sum_{f\in\mathcal F_{\mathrm{LF}}}\widetilde P(f)$.
Formally, we have $\boldsymbol{\phi}^{\text{inst}}_n=\big[r_{\mathrm{LF}}^{(n)},\ H^{(n)}\big]\in\mathbb{R}^{2}$.

\paragraph{MLPs training.} 
For \textit{intra-step} and \textit{inter-step} levels, we maintain a sliding window tracking the most recent $k$ steps, which shifts to discard the earliest step upon detecting a new step separator (``\texttt{\textbackslash n \textbackslash n}'') during token generation. Notably, we optimize feature updates within this variable-length window to achieve an approximate $O(1)$ time complexity. For the \textit{instance} level, consistent with Section \ref{sec:dive-1}, we directly utilize the initial $K$ steps. Further details on MLPs training can be found in Appendix \ref{sec:app-mlp}.

\subsection{Trigger Training}
\label{sec:method-2}
Leveraging the trained MLPs to localize failure modes within the sliding window, we introduce special tokens as triggers to enforce controllable self-correction. Specifically, for the intra-step and inter-step levels, we employ dataset reconstruction followed by SFT to condition the model to execute diagnose-then-correct behaviors upon encountering these tokens. In contrast, for the instance level, we adopt a direct, training-free strategy, which will be elaborated upon in the next subsection.
\paragraph{Data Reconstruction.} To enable the model to learn the mechanism where special tokens serve as triggers to elicit specific behavioral patterns, we reconstruct the raw dataset $\mathcal{D}_{\mathrm{raw}}=\{(\mathbf{u}^{(i)},\,\mathbf{v}^{(i)})\}_{i=1}^{|\mathcal{D}_{\mathrm{raw}}|}$, where $\mathbf{u}$ and $\mathbf{v}$ denote the input and output, respectively. We first apply a segmentation operation $\mathrm{Seg}(\cdot)$ to divide $\mathbf{v}$ into step-level subsequences $\mathbf{v}=\langle \pi_1\|\cdots\|\pi_K\rangle$. Let $\mathcal{J}$ denote the set of critical steps to be perturbed. We define the failure mode levels as $\ell\in\{\mathrm{intra},\mathrm{inter}\}$ and introduce a rewriting operation $\Psi_{\ell}(\cdot)$, yielding:
\vspace{-0.3em}
\begin{equation}
\label{eq:rewrite}
\widetilde{\pi}_j^{(\ell)}=\Psi_{\ell}\!\big(\mathbf{u},\,\langle \pi_1,\ldots,\pi_K\rangle,\,j\big),\, j\in\mathcal{J}.
\end{equation}
We introduce a set of mode-specific triggers $\mathcal{U}=\{\langle \upsilon_{\mathrm{intra}}\rangle,\ \langle \upsilon_{\mathrm{inter}}\rangle\}$, which are inserted immediately following $\Psi_{\ell}(\cdot)$. 
For each level $\ell$, we associate a diagnose-then-correct template $\mathcal{B}_{\ell}=\langle \mathbf{p}_{\ell},\,\mathbf{d}_{\ell},\,\mathbf{c}_{\ell}\rangle$, where $\mathbf{p}_{\ell}$ serves as a prompt signaling the detected failure mode, followed by the diagnostic ($\mathbf{d}_{\ell}$) and corrective ($\mathbf{c}_{\ell}$) behaviors.
Consequently, we obtain the reconstructed dataset $\widehat{\mathcal{D}}=\{(\mathbf{u}^{(i)},\,\widehat{\mathbf{v}}^{(i)})\}_{i=1}^{|\widehat{\mathcal{D}}|}$, where the constructed output is: 
\vspace{-0.3em}
\begin{equation}
\label{eq:reconstruct}
\widehat{\mathbf{v}}^{(\ell,j)}
=\big\langle \pi_{<j},\ \widetilde{\pi}_j^{(\ell)},\ \langle \upsilon_{\ell}\rangle,\ \mathcal{B}_{\ell},\ \pi_{>j}\big\rangle.
\end{equation}
See Appendix \ref{sec:prompt} for the detailed reconstructed data.

\paragraph{SFT.}
Our training objective maximizes the following factorized likelihood over step blocks:
\begin{align}
P_\theta(\pi_{<j}\mid \mathbf{u})
&\cdot
P_\theta(\widetilde{\pi}_j^{(\ell)}\mid \mathbf{u},\pi_{<j})
\notag\\
&\cdot
P_\theta(\mathcal{B}_\ell\mid \mathbf{u},\pi_{<j},\widetilde{\pi}_j^{(\ell)},\langle \upsilon_\ell\rangle)
\label{eq:sft-block}\\
&\cdot
P_\theta(\pi_{>j}\mid \mathbf{u},\pi_{<j},\widetilde{\pi}_j^{(\ell)},\langle \upsilon_\ell\rangle,\mathcal{B}_\ell),
\notag
\end{align}
where $\theta$ denotes the parameters.
Training follows standard next-token prediction with a label mask:
\begin{equation}
\small
\mathcal{L}_{\mathrm{SFT}}(\theta)
=
-\mathbb{E}_{(\mathbf{u},\widehat{\mathbf{v}})\sim\widehat{\mathcal{D}}}
\sum_{t\in\mathcal{I}(\mathbf{u},\widehat{\mathbf{v}})}
\log P_\theta\!\left(y_t \mid \mathbf{u}, y_{<t}\right),
\end{equation}
where $\{y_t\}$ enumerates tokens in the concatenated sequence $\langle \mathbf{u},\widehat{\mathbf{v}}\rangle$ and
$\mathcal{I}(\mathbf{u},\widehat{\mathbf{v}})$ indexes the tokens that contribute to the loss.
Notably, we mask $\mathbf{u}$ and $\langle \upsilon_\ell\rangle$ during training, ensuring the trigger operates solely as an inference-time control signal.
\subsection{Online Monitoring}
\label{sec:method-3}
During inference, we run our trained lightweight MLPs in parallel to conduct online monitoring of the \mon over designated token sequences. For \textit{intra-step} and \textit{inter-step} levels, this sequence corresponds to the sliding window defined in Section \ref{sec:method-1}. Upon detecting fluctuation patterns characteristic of failure modes at level $\ell \in \{\mathrm{intra}, \mathrm{inter}\}$, we intervene by forcibly inserting the trigger token $\langle \upsilon_\ell \rangle$. Formally, letting $x$ denote the current decoding prefix with length $\tau=|x|$, we impose a hard constraint on the next token $y_{\tau+1} := \langle \upsilon_\ell \rangle$, and continue autoregressive decoding as:
\vspace{-0.3em}
\begin{equation}
    y_{\tau+k} \sim P_\theta(\cdot \mid x, \langle \upsilon_\ell \rangle, y_{\tau+1:\tau+k-1}), \, k \ge 2.
\end{equation} 
This intervention activates the diagnose-then-correct behavior pattern learned during SFT. In contrast, for the \textit{instance} level, the monitored sequence comprises the initial $K$ steps of the reasoning process (consistent with Section \ref{sec:method-1}). Upon detecting the corresponding fluctuations, we directly insert the prompt (\texttt{``Okay, I have finished thinking.''}), thereby triggering a transition to the \textit{NoThinking} mode~\citep{nothinking}.

\begin{table*}[!t]
\centering

\renewcommand\tabcolsep{5.3pt}
\renewcommand\arraystretch{1.1}

\resizebox{\linewidth}{!}{
\begin{tabular}{c|l|cc|cc|cc|cc|cc}
\Xhline{1.2pt}
\rowcolor{MorandiHeader}
\textbf{Model} & \textbf{Method} &
\multicolumn{2}{c|}{\textbf{AIME25}} &
\multicolumn{2}{c|}{\textbf{MATH500}} &
\multicolumn{2}{c|}{\textbf{GSM8K}} &
\multicolumn{2}{c|}{\textbf{GPQA-Diamond}} &
\multicolumn{2}{c}{\textbf{LiveCodeBench}} \\
\rowcolor{MorandiHeader}
& &
\textbf{Pass@1 $\uparrow$} & \textbf{Token $\downarrow$} &
\textbf{Pass@1 $\uparrow$} & \textbf{Token $\downarrow$} &
\textbf{Pass@1 $\uparrow$} & \textbf{Token $\downarrow$} &
\textbf{Pass@1 $\uparrow$} & \textbf{Token $\downarrow$} &
\textbf{Pass@1 $\uparrow$} & \textbf{Token $\downarrow$} \\
\Xhline{1.2pt}

\multirow{10}{*}{\rotatebox{90}{DeepSeek-R1-Distill-Qwen-7B}}
& Vanilla &
  43.3 & 11454 &
  91.8 & 2887 &
  92.4 & 442 &
  49.2 & 8016 &
  36.9 & 9072 \\
& \gcell{DAST} &
  \gcell{41.1} & \gcell{7904} &
  \gcell{91.6} & \gcell{1330} &
  \gcell{91.8} & \gcell{301} &
  \gcell{48.8} & \gcell{3635} &
  \gcell{36.8} & \gcell{\textbf{4280}} \\
& Think or Not &
  38.9 & \textbf{4760} &
  92.0 & \textbf{1103} &
  92.9 & \textbf{264} &
  47.0 & 3390 &
  36.4 & 7397 \\
& \gcell{AlphaOne} &
  \gcell{44.4} & \gcell{8921} &
  \gcell{92.5} & \gcell{3791} &
  \gcell{93.4} & \gcell{459} &
  \gcell{\underline{50.5}} & \gcell{8591} &
  \gcell{37.5} & \gcell{8177} \\
& RL + LP &
  43.3 & 6255 &
  93.4 & 1322 &
  92.5 & 291 &
  50.3 & \underline{3209} &
  36.2 & 6293 \\
& \gcell{GRPO} &
  \gcell{45.6} & \gcell{12006} &
  \gcell{93.9} & \gcell{2873} &
  \gcell{92.1} & \gcell{275} &
  \gcell{49.8} & \gcell{8890} &
  \gcell{\underline{39.2}} & \gcell{10084} \\
& S-GRPO &
  44.4 & 5794 &
  93.2 & 1204 &
  \underline{93.7} & 297 &
  49.5 & \textbf{3107} &
  36.9 & 5449 \\
& \gcell{DAPO} &
  \gcell{\underline{46.7}} & \gcell{12307} &
  \gcell{\underline{94.2}} & \gcell{3033} &
  \gcell{92.9} & \gcell{301} &
  \gcell{50.2} & \gcell{9102} &
  \gcell{\textbf{40.1}} & \gcell{11407} \\
& \ourmethod &
  \textbf{47.8} & \underline{5446} &
  \textbf{95.1} & \underline{1197} &
  \textbf{94.0} & \underline{271} &
  \textbf{51.3} & 4011 &
  39.0 & \underline{5311} \\
& \dcell{$\Delta$} &
  \dcell{\darkred{\textbf{4.5}}} & \dcell{\darkblue{\textbf{52.5\%}}} &
  \dcell{\darkred{\textbf{3.3}}} & \dcell{\darkblue{\textbf{58.5\%}}} &
  \dcell{\darkred{\textbf{1.6}}} & \dcell{\darkblue{\textbf{38.7\%}}} &
  \dcell{\darkred{\textbf{2.1}}} & \dcell{\darkblue{\textbf{50.0\%}}} &
  \dcell{\darkred{\textbf{2.1}}} & \dcell{\darkblue{\textbf{41.5\%}}} \\
\hline

\multirow{10}{*}{\rotatebox{90}{DeepSeek-R1-Distill-Qwen-32B}}
& Vanilla &
  58.9 & 8906 &
  93.3 & 2337 &
  93.9 & 438 &
  60.8 & 6027 &
  55.8 & 7125 \\
& \gcell{DAST} &
  \gcell{54.4} & \gcell{6647} &
  \gcell{93.5} & \gcell{1421} &
  \gcell{93.9} & \gcell{266} &
  \gcell{60.3} & \gcell{4048} &
  \gcell{56.0} & \gcell{\textbf{3419}} \\
& Think or Not &
  58.9 & \underline{4711} &
  92.9 & 1410 &
  93.3 & \underline{247} &
  60.4 & 3706 &
  54.8 & 5216 \\
& \gcell{AlphaOne} &
  \gcell{61.1} & \gcell{8994} &
  \gcell{94.4} & \gcell{3037} &
  \gcell{94.1} & \gcell{433} &
  \gcell{61.3} & \gcell{6771} &
  \gcell{\underline{56.1}} & \gcell{8223} \\
& RL + LP &
  \underline{62.2} & 6124 &
  93.1 & \underline{1379} &
  \textbf{94.7} & \textbf{239} &
  60.4 & 3596 &
  55.9 & 4729 \\
& \gcell{GRPO} &
  \gcell{\underline{62.2}} & \gcell{8990} &
  \gcell{93.4} & \gcell{3012} &
  \gcell{\underline{94.4}} & \gcell{420} &
  \gcell{60.4} & \gcell{7123} &
  \gcell{\underline{56.1}} & \gcell{8974} \\
& S-GRPO &
  60.0 & 5334 &
  94.2 & 1556 &
  \textbf{94.7} & 269 &
  61.3 & \textbf{3119} &
  54.9 & \underline{4404} \\
& \gcell{DAPO} &
  \gcell{\textbf{63.3}} & \gcell{9093} &
  \gcell{\underline{94.6}} & \gcell{3551} &
  \gcell{94.3} & \gcell{443} &
  \gcell{\underline{61.4}} & \gcell{7390} &
  \gcell{55.9} & \gcell{8700} \\
& \ourmethod &
  \textbf{63.3} & \textbf{4702} &
  \textbf{96.3} & \textbf{1318} &
  94.2 & \underline{247} &
  \textbf{62.3} & \underline{3498} &
  \textbf{57.2} & 4689 \\
& \dcell{$\Delta$} &
  \dcell{\darkred{\textbf{4.4}}} & \dcell{\darkblue{\textbf{47.2\%}}} &
  \dcell{\darkred{\textbf{3.0}}} & \dcell{\darkblue{\textbf{43.6\%}}} &
  \dcell{\darkred{\textbf{0.3}}} & \dcell{\darkblue{\textbf{43.6\%}}} &
  \dcell{\darkred{\textbf{1.5}}} & \dcell{\darkblue{\textbf{42.0\%}}} &
  \dcell{\darkred{\textbf{1.4}}} & \dcell{\darkblue{\textbf{34.2\%}}} \\
\hline

\multirow{10}{*}{\rotatebox{90}{Qwen3-8B-thinking}}
& Vanilla &
  61.1 & 12490 &
  95.9 & 4486 &
  94.1 & 1573 &
  58.8 & 6638 &
  54.5 & 8549 \\
& \gcell{DAST} &
  \gcell{60.0} & \gcell{\textbf{6055}} &
  \gcell{96.0} & \gcell{\textbf{2158}} &
  \gcell{94.1} & \gcell{\textbf{483}} &
  \gcell{58.4} & \gcell{\textbf{3007}} &
  \gcell{54.7} & \gcell{\textbf{4987}} \\
& Think or Not &
  62.2 & \underline{6674} &
  96.3 & 2408 &
  94.2 & 599 &
  59.1 & 3678 &
  53.9 & 6009 \\
& \gcell{AlphaOne} &
  \gcell{\underline{63.3}} & \gcell{8607} &
  \gcell{96.7} & \gcell{4311} &
  \gcell{94.5} & \gcell{1095} &
  \gcell{\underline{59.8}} & \gcell{6002} &
  \gcell{\underline{56.2}} & \gcell{8607} \\
& RL + LP &
  58.9 & 7639 &
  96.6 & 2692 &
  94.1 & 823 &
  59.4 & 3139 &
  55.9 & 6792 \\
& \gcell{GRPO} &
  \gcell{62.2} & \gcell{11098} &
  \gcell{96.7} & \gcell{4225} &
  \gcell{94.4} & \gcell{1373} &
  \gcell{59.1} & \gcell{7472} &
  \gcell{55.3} & \gcell{9073} \\
& S-GRPO &
  61.1 & 7331 &
  96.5 & 2576 &
  \underline{94.6} & 775 &
  59.6 & \underline{3046} &
  54.3 & \underline{5998} \\
& \gcell{DAPO} &
  \gcell{62.2} & \gcell{12004} &
  \gcell{\underline{96.9}} & \gcell{4609} &
  \gcell{\textbf{94.7}} & \gcell{1553} &
  \gcell{59.4} & \gcell{8008} &
  \gcell{\textbf{56.3}} & \gcell{9937} \\
& \ourmethod &
  \textbf{64.4} & 7279 &
  \textbf{97.2} & \underline{2402} &
  \underline{94.6} & \underline{588} &
  \textbf{61.4} & 5153 &
  \underline{56.2} & 6002 \\
& \dcell{$\Delta$} &
  \dcell{\darkred{\textbf{3.3}}} & \dcell{\darkblue{\textbf{41.7\%}}} &
  \dcell{\darkred{\textbf{1.3}}} & \dcell{\darkblue{\textbf{46.5\%}}} &
  \dcell{\darkred{\textbf{0.5}}} & \dcell{\darkblue{\textbf{62.6\%}}} &
  \dcell{\darkred{\textbf{2.6}}} & \dcell{\darkblue{\textbf{22.4\%}}} &
  \dcell{\darkred{\textbf{1.7}}} & \dcell{\darkblue{\textbf{29.8\%}}} \\
\hline

\multirow{10}{*}{\rotatebox{90}{Qwen3-32B-thinking}}
& Vanilla &
  68.9 & 11589 &
  96.8 & 4318 &
  94.3 & 1435 &
  65.3 & 5475 &
  64.7 & 8725 \\
& \gcell{DAST} &
  \gcell{68.9} & \gcell{6504} &
  \gcell{96.8} & \gcell{2417} &
  \gcell{94.2} & \gcell{\underline{609}} &
  \gcell{65.7} & \gcell{\underline{2918}} &
  \gcell{64.6} & \gcell{\textbf{5108}} \\
& Think or Not &
  65.6 & 6772 &
  97.1 & \textbf{2179} &
  94.5 & 641 &
  64.5 & \textbf{1713} &
  62.9 & 7337 \\
& \gcell{AlphaOne} &
  \gcell{71.1} & \gcell{8569} &
  \gcell{\underline{97.8}} & \gcell{3170} &
  \gcell{94.4} & \gcell{1090} &
  \gcell{\underline{66.8}} & \gcell{5591} &
  \gcell{\underline{66.1}} & \gcell{9003} \\
& RL + LP &
  66.7 & 6854 &
  97.2 & 2701 &
  \textbf{94.7} & 822 &
  66.0 & 4362 &
  65.1 & 7382 \\
& \gcell{GRPO} &
  \gcell{70.0} & \gcell{12487} &
  \gcell{97.1} & \gcell{4641} &
  \gcell{94.5} & \gcell{1558} &
  \gcell{66.2} & \gcell{6151} &
  \gcell{65.4} & \gcell{9546} \\
& S-GRPO &
  71.1 & \underline{6389} &
  97.3 & 2566 &
  \underline{94.6} & 809 &
  66.5 & 4151 &
  63.9 & 7560 \\
& \gcell{DAPO} &
  \gcell{\underline{72.2}} & \gcell{12403} &
  \gcell{97.7} & \gcell{5152} &
  \gcell{\textbf{94.7}} & \gcell{1970} &
  \gcell{65.8} & \gcell{6099} &
  \gcell{65.9} & \gcell{10124} \\
& \ourmethod &
  \textbf{73.3} & \textbf{6308} &
  \textbf{97.9} & \underline{2398} &
  \underline{94.6} & \textbf{574} &
  \textbf{67.3} & 4043 &
  \textbf{66.9} & \underline{7012} \\
& \dcell{$\Delta$} &
  \dcell{\darkred{\textbf{4.4}}} & \dcell{\darkblue{\textbf{45.6\%}}} &
  \dcell{\darkred{\textbf{1.1}}} & \dcell{\darkblue{\textbf{44.5\%}}} &
  \dcell{\darkred{\textbf{0.3}}} & \dcell{\darkblue{\textbf{60.0\%}}} &
  \dcell{\darkred{\textbf{2.0}}} & \dcell{\darkblue{\textbf{26.2\%}}} &
  \dcell{\darkred{\textbf{2.2}}} & \dcell{\darkblue{\textbf{19.6\%}}} \\
\hline

\multirow{10}{*}{\rotatebox{90}{DeepSeek-R1-Distill-Llama-8B}}
& Vanilla &
  28.9 & 11548 &
  86.2 & 3635 &
  92.3 & 606 &
  46.3 & 8341 &
  38.9 & 9588 \\
& \gcell{DAST} &
  \gcell{32.2} & \gcell{8438} &
  \gcell{87.0} & \gcell{2458} &
  \gcell{91.9} & \gcell{\underline{388}} &
  \gcell{46.1} & \gcell{4410} &
  \gcell{38.4} & \gcell{4091} \\
& Think or Not &
  30.0 & 7158 &
  87.4 & \underline{1954} &
  92.5 & \textbf{257} &
  46.8 & 3729 &
  38.4 & 6916 \\
& \gcell{AlphaOne} &
  \gcell{\underline{34.4}} & \gcell{9005} &
  \gcell{89.1} & \gcell{3804} &
  \gcell{93.1} & \gcell{598} &
  \gcell{\underline{47.6}} & \gcell{8569} &
  \gcell{41.2} & \gcell{8421} \\
& RL + LP &
  30.0 & \underline{5897} &
  89.4 & 2290 &
  92.3 & 446 &
  45.3 & \textbf{3299} &
  39.8 & \underline{3491} \\
& \gcell{GRPO} &
  \gcell{30.0} & \gcell{11987} &
  \gcell{89.6} & \gcell{3309} &
  \gcell{92.9} & \gcell{571} &
  \gcell{46.6} & \gcell{8783} &
  \gcell{40.0} & \gcell{10123} \\
& S-GRPO &
  31.1 & \textbf{5426} &
  89.1 & 2195 &
  93.2 & 432 &
  47.0 & \underline{3624} &
  39.2 & \textbf{3349} \\
& \gcell{DAPO} &
  \gcell{\underline{34.4}} & \gcell{12399} &
  \gcell{\underline{90.0}} & \gcell{3732} &
  \gcell{\textbf{93.5}} & \gcell{4004} &
  \gcell{\textbf{47.8}} & \gcell{9110} &
  \gcell{\underline{41.7}} & \gcell{11394} \\
& \ourmethod &
  \textbf{36.7} & 6076 &
  \textbf{91.2} & \textbf{1887} &
  \underline{93.4} & 419 &
  \textbf{47.8} & 4539 &
  \textbf{42.5} & 3522 \\
& \dcell{$\Delta$} &
  \dcell{\darkred{\textbf{7.8}}} & \dcell{\darkblue{\textbf{47.4\%}}} &
  \dcell{\darkred{\textbf{5.0}}} & \dcell{\darkblue{\textbf{48.1\%}}} &
  \dcell{\darkred{\textbf{1.1}}} & \dcell{\darkblue{\textbf{30.9\%}}} &
  \dcell{\darkred{\textbf{1.5}}} & \dcell{\darkblue{\textbf{45.6\%}}} &
  \dcell{\darkred{\textbf{3.6}}} & \dcell{\darkblue{\textbf{63.3\%}}} \\
\hline

\multirow{10}{*}{\rotatebox{90}{DeepSeek-R1-Distill-Llama-70B}}
& Vanilla &
  47.8 & 8909 &
  94.1 & 2433 &
  94.0 & 432 &
  64.5 & 5881 &
  55.6 & 7258 \\
& \gcell{DAST} &
  \gcell{45.6} & \gcell{5932} &
  \gcell{94.2} & \gcell{1563} &
  \gcell{93.9} & \gcell{\underline{239}} &
  \gcell{63.5} & \gcell{4026} &
  \gcell{56.3} & \gcell{3975} \\
& Think or Not &
  46.7 & \textbf{4414} &
  94.4 & 1360 &
  93.7 & 241 &
  64.6 & \textbf{3544} &
  52.9 & 5668 \\
& \gcell{AlphaOne} &
  \gcell{50.0} & \gcell{8689} &
  \gcell{95.3} & \gcell{2009} &
  \gcell{94.2} & \gcell{427} &
  \gcell{65.3} & \gcell{4322} &
  \gcell{56.8} & \gcell{\underline{3742}} \\
& RL + LP &
  45.6 & 5967 &
  93.8 & \underline{1149} &
  94.4 & 255 &
  65.0 & 4101 &
  56.2 & 5209 \\
& \gcell{GRPO} &
  \gcell{51.1} & \gcell{8715} &
  \gcell{94.5} & \gcell{3167} &
  \gcell{\underline{94.7}} & \gcell{473} &
  \gcell{64.8} & \gcell{6274} &
  \gcell{56.4} & \gcell{7561} \\
& S-GRPO &
  48.9 & \underline{5896} &
  95.1 & 1252 &
  94.5 & 249 &
  65.2 & \underline{4001} &
  54.8 & 4689 \\
& \gcell{DAPO} &
  \gcell{\textbf{53.3}} & \gcell{9054} &
  \gcell{\underline{95.4}} & \gcell{3776} &
  \gcell{\textbf{94.9}} & \gcell{509} &
  \gcell{\underline{65.7}} & \gcell{6968} &
  \gcell{\underline{57.3}} & \gcell{8049} \\
& \ourmethod &
  \underline{52.2} & 5907 &
  \textbf{96.8} & \textbf{1125} &
  \textbf{94.9} & \textbf{208} &
  \textbf{66.8} & 4084 &
  \textbf{57.5} & \textbf{3308} \\
& \dcell{$\Delta$} &
  \dcell{\darkred{\textbf{4.4}}} & \dcell{\darkblue{\textbf{33.7\%}}} &
  \dcell{\darkred{\textbf{2.7}}} & \dcell{\darkblue{\textbf{53.8\%}}} &
  \dcell{\darkred{\textbf{0.9}}} & \dcell{\darkblue{\textbf{51.9\%}}} &
  \dcell{\darkred{\textbf{2.3}}} & \dcell{\darkblue{\textbf{30.6\%}}} &
  \dcell{\darkred{\textbf{1.9}}} & \dcell{\darkblue{\textbf{54.4\%}}} \\
\Xhline{1.2pt}
\end{tabular}
}
\vspace{-0.8em}
\caption{Main results. Best results are highlighted in \textbf{bold}, with runners-up \underline{underlined}.}
\label{tab:main}
\vspace{-1.0em}
\end{table*}

\vspace{-0.7em}
\section{Experiments}
\vspace{-0.5em}
In this section, we conduct extensive experiments to address the following research questions: ($\boldsymbol{\mathcal{RQ}1}$) Can \ourmethod achieve dual superiority in terms of performance and cost? ($\boldsymbol{\mathcal{RQ}2}$) Does \ourmethod demonstrate effective scalability? ($\boldsymbol{\mathcal{RQ}3}$) Specifically, how does \ourmethod realize improvements during the reasoning process? ($\boldsymbol{\mathcal{RQ}4}$) What are the distinct contributions of the MLPs within \ourmethod? 
\vspace{-0.5em}
\subsection{Experimental Setup}
\paragraph{Backbones.} We conduct experiments using representative open-source LRMs with diverse architectures from different families. To demonstrate the effectiveness and scalability of \ourmethod, the selected models span a wide range of scales, \textit{\underline{from 7B to 70B}}: \textbf{I) \textit{Qwen family}}, DeepSeek-R1-Distill-Qwen-7B, DeepSeek-R1-Distill-Qwen-32B~\citep{r1}, Qwen3-8B-thinking, and Qwen3-32B-thinking~\citep{qwen3}; and \textbf{II) \textit{Llama family}}, DeepSeek-R1-Distill-Llama-8B and DeepSeek-R1-Distill-Llama-70B.
\vspace{-0.4em}
\paragraph{Baselines.} We compare \ourmethod against a comprehensive set of representative baselines, categorized into three groups: \textbf{I) \textit{Vanilla Model}}, referring to the original backbone LRM itself; \textbf{II) \textit{Training-free}}, including Think or Not~\citep{ton}, AlphaOne~\citep{alphaone}, and Self-Consistency~\citep{sc}; and \textbf{III) \textit{RL-based}}, including DAST~\citep{dast}, RL + Length Penalty~\citep{rl+lp}, GRPO~\citep{grpo}, S-GRPO~\citep{sgrpo} and DAPO~\citep{dapo}.
\vspace{-0.4em}
\paragraph{Benchmarks.} We conduct extensive evaluations of \ourmethod on five benchmarks spanning three complex reasoning domains: \textbf{I) \textit{Mathematical Reasoning}}, including GSM8K~\citep{gsm8k}, MATH500~\citep{math500}, AIME24 and AIME25~\citep{aime}; \textbf{II) \textit{Scientific Reasoning}}, including GPQA-Diamond~\citep{gpqa}; and \textbf{III) \textit{Code Reasoning}}, including LiveCodeBench~\citep{livecodebench}.

\begin{figure*}[!t]
  \centering
  \includegraphics[width=\linewidth]{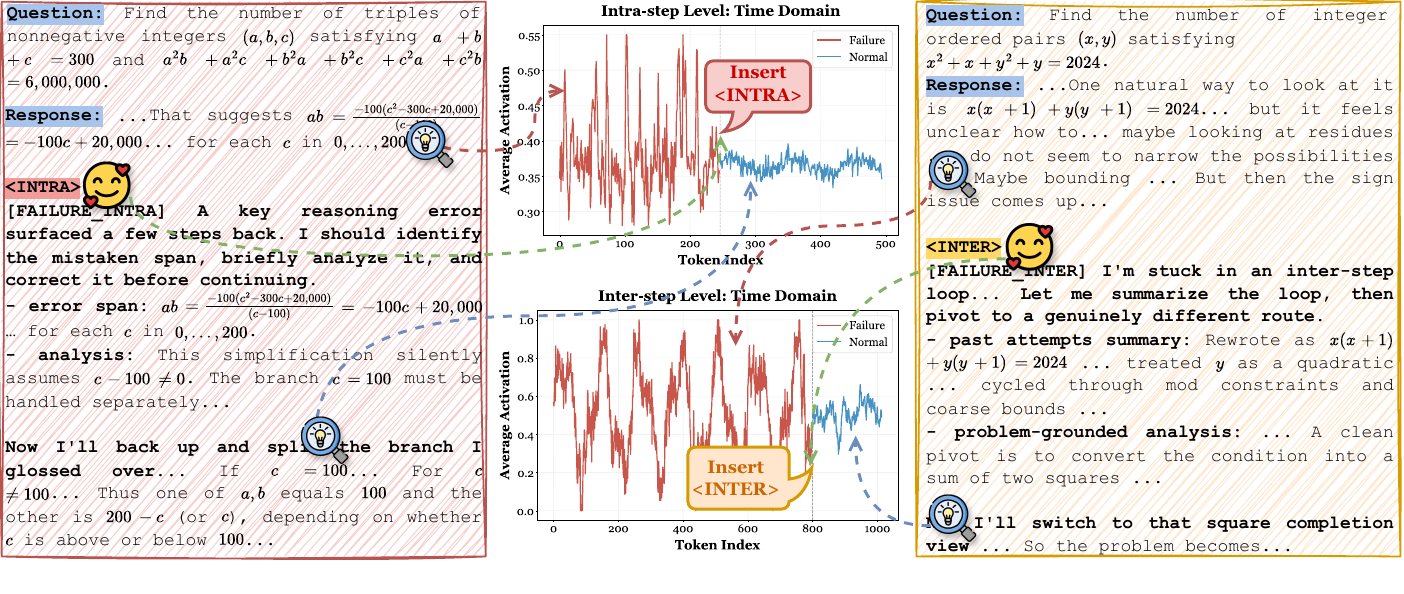}
  \vspace{-2.5em}
  \caption{Case studies of \ourmethod.}
   \label{fig:case}
   \vspace{-1.5em}
\end{figure*}

\begin{figure}[!t]
  \centering
  \includegraphics[width=\linewidth]{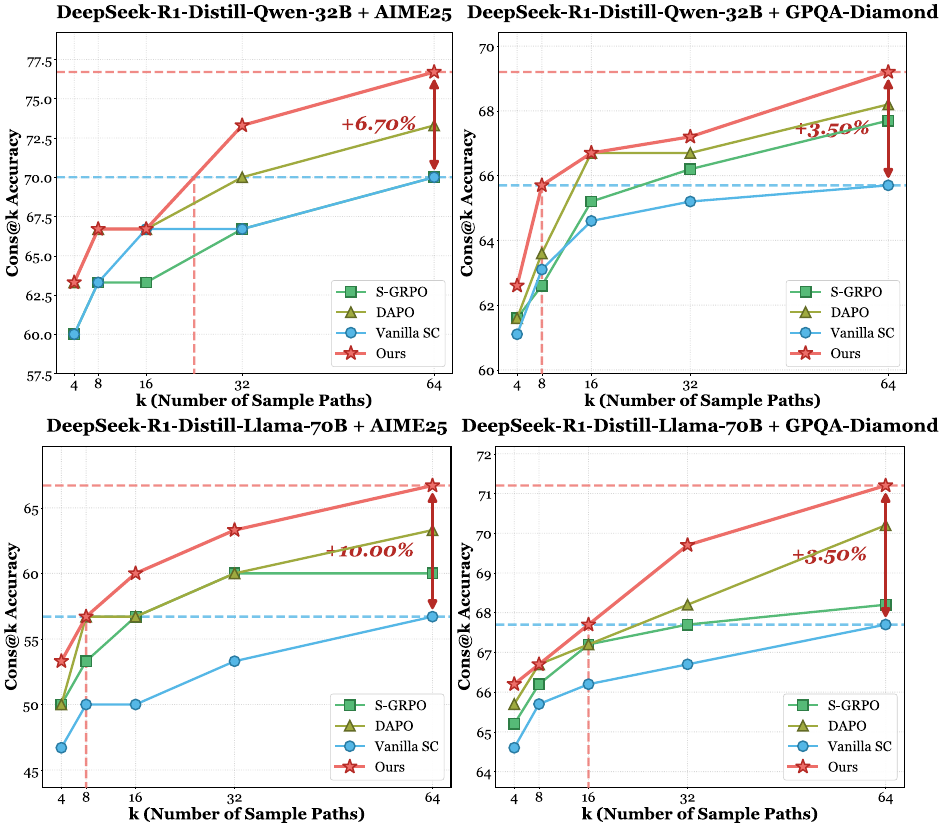}
  \vspace{-2em}
  \caption{Test-time scalability under self-consistency.}
   \label{fig:consk}
   \vspace{-1.5em}
\end{figure}
\vspace{-0.5em}
\paragraph{Implementation details.} The sliding window length $k$ for intra-step and inter-step levels, as well as the initial step count $K$ for the instance level, are selected from the set $\{2, 4, 8\}$. The failure mode detector (Section \ref{sec:dive-2}), the rewriting operation $\Psi_{\ell}(\cdot)$ (Equation \ref{eq:rewrite}), and the diagnose-then-correct operator $\mathcal{B}_{\ell}$ (Equation \ref{eq:reconstruct}), are consistently implemented by \texttt{gpt-5}. For the segmentation operator $\mathrm{Seg}(\cdot)$, we implement step-level splitting using paragraph delimiters (``\texttt{\textbackslash n\textbackslash n}’’). We use BS-17k~\citep{bs17k} as the raw corpus for reconstruction and conduct full-parameter fine-tuning. Following~\citep{r1}, we use a sampling temperature of $0.6$ and top-$p$ of $0.95$. We report Pass@1 by averaging results over $5$ repeated runs. Detailed technical details are provided in Appendix \ref{sec:app-tech}. All prompts utilized are detailed in Appendix \ref{sec:prompt}.

\vspace{-0.5em}
\subsection{Main Results (\texorpdfstring{$\boldsymbol{\mathcal{RQ}1}$}{})}
To address $\mathcal{RQ}1$, we conduct extensive evaluations on six benchmarks, comparing \ourmethod against nine competitive baseline methods. Experimental results are reported in Table \ref{tab:main} and \ref{tab:aime24}.
\vspace{-0.5em}
\paragraph{Obs.\ding{182} \ourmethod achieves dual superiority in both performance and token cost.} Overall, compared to the vanilla model, \ourmethod delivers performance gains ranging from $0.3 \sim 7.8$ ($3.2\%\sim27.0\%$), while simultaneously reducing token consumption by $19.6\%\sim63.3\%$. Notably, on DeepSeek-R1-Distill-Qwen-7B + MATH500, \ourmethod improves accuracy by $5.0\uparrow$ and cuts token usage by $48.1\%\downarrow$, outperforming all baselines in both metrics. Even against DAPO, the most competitive baseline, \ourmethod achieves a $0.9\uparrow$ performance lead with a substantial $60.5\%\downarrow$ reduction in token consumption. 
\vspace{-0.3em}
\paragraph{Obs.\ding{183} \ourmethod exhibits robust cross-task generalization.} On DeepSeek-R1-Distill-Qwen-32B, while DAPO performs adequately on math and science tasks, it struggles on LiveCodeBench (showing a negligible $0.1\uparrow$ improvement with a $22.1\%\uparrow$ cost overhead). In stark contrast, \ourmethod achieves a $1.4\uparrow$ performance gain with a $34.2\%\downarrow$ cost reduction on this benchmark, while securing state-of-the-art results across AIME25, MATH500, and GPQA-Diamond.
\vspace{-0.5em}
\subsection{Scalability Analysis (\texorpdfstring{$\boldsymbol{\mathcal{RQ}2}$}{})}
\vspace{-0.4em}
We examine scalability across two dimensions: \textit{model scale} and \textit{test-time}. To evaluate test-time scalability, we benchmark \ourmethod against baselines using Cons@$k$, where $k \in \{4, 8, 16, 32, 64\}$, as illustrated in Figure \ref{fig:consk}.
\vspace{-0.5em}
\paragraph{Obs.\ding{184} \ourmethod demonstrates robust scalability across both model scale and test-time dimensions.} Regarding model scale, Table \ref{tab:main} reveals that on DeepSeek-R1-Distill-Llama series + AIME25, while DAST suffers a performance decline, shifting from a $3.3\uparrow$ gain at 8B to a $2.2\downarrow$ drop at 70B relative to the vanilla, \ourmethod maintains consistent gains of $7.8\uparrow$ and $4.4\uparrow$ at 8B and 70B, respectively. Regarding test-time scalability, Figure \ref{fig:consk} illustrates that \ourmethod achieves superior Cons@$k$ performance across nearly all $k$. On DeepSeek-R1-Distill-Qwen-32B + GPQA-Diamond, \ourmethod at $k=8$ matches the performance of the vanilla at $k=64$. Furthermore, at $k=64$, \ourmethod outperforms the vanilla by $3.5\%\sim10.0\%$, underscoring its scalability.

\vspace{-0.5em}
\subsection{Case Study (\texorpdfstring{$\boldsymbol{\mathcal{RQ}3}$}{})}
\vspace{-0.3em}
\paragraph{Obs.\ding{185} \ourmethod demonstrates explainability and controllability.} As visualized in Figure \ref{fig:case} for intra- and inter-step levels
, \ourmethod successfully detects failure mode fluctuations during inference and triggers the diagnose-then-correct behavior via special token insertion. Crucially, we observe corresponding shifts in \mon activation dynamics post-intervention. These findings collectively substantiate the explainability and controllability.

\vspace{-0.5em}
\subsection{Framework Analysis (\texorpdfstring{$\boldsymbol{\mathcal{RQ}4}$}{})}
To address $\mathcal{RQ}4$, we evaluate four variants by ablating specific components: \textbf{(1) \textit{w/o}} $\text{MLP}_{\mathrm{intra}}$, \textbf{(2) \textit{w/o}} $\text{MLP}_{\mathrm{inter}}$, \textbf{(3) \textit{w/o}} $\text{MLP}_{\mathrm{inst}}$, and \textbf{(4) \textit{w/o}} $\text{MLP}_{\mathrm{all}}$. As shown in Table \ref{tab:ab}, the removal of $\text{MLP}_{\mathrm{intra}}$ induces the most significant drop, corroborating the predominance of intra-step failures. While removing $\text{MLP}_{\mathrm{inter}}$ yields minor deficits, ablating $\text{MLP}_{\mathrm{inst}}$ triggers a drastic surge in token consumption, validating the necessity of difficulty-aware strategies. See Appendix \ref{sec:app-sens} for sensitivity analysis.

\begin{table}[!t]
\centering

\resizebox{\linewidth}{!}{
\begin{tabular}{c|l|cc|cc}
\Xhline{1.2pt}
\rowcolor{MorandiHeader}
\textbf{Model} & \textbf{Method} &
\multicolumn{2}{c|}{\textbf{MATH500}} &
\multicolumn{2}{c}{\textbf{GPQA}} \\
\rowcolor{MorandiHeader}
& &
\textbf{Pass@1 $\uparrow$} & \textbf{Token $\downarrow$} &
\textbf{Pass@1 $\uparrow$} & \textbf{Token $\downarrow$} \\
\Xhline{1.2pt}

\multirow{5}{*}{\rotatebox{90}{\shortstack{DS-R1-Distill\\Qwen-32B}}}
& \textbf{\textit{w/o}} $\text{MLP}_\text{all}$ & 
  93.5 & 2469 &
  63.9 & 6010 \\
& \gcell{\textbf{\textit{w/o}} $\text{MLP}_\text{intra}$} & 
  \gcell{94.5} & \gcell{1412} &
  \gcell{64.4} & \gcell{3441} \\
& \textbf{\textit{w/o}} $\text{MLP}_\text{inter}$ & 
  94.7 & 1434 &
  64.9 & 3219 \\
& \gcell{\textbf{\textit{w/o}} $\text{MLP}_\text{inst}$} & 
  \gcell{95.8} & \gcell{2389} &
  \gcell{65.4} & \gcell{6118} \\
& \ourmethod & 
  96.3 & 1318 &
 66.2 & 3274 \\
\hline

\multirow{5}{*}{\rotatebox{90}{{\shortstack{DS-R1-Distill\\Llama-70B}}}}
& \textbf{\textit{w/o}} $\text{MLP}_\text{all}$ & 
  94.2 & 2523 &
  65.9 & 5721 \\
& \gcell{\textbf{\textit{w/o}} $\text{MLP}_\text{intra}$} & 
  \gcell{95.9} & \gcell{1293} &
  \gcell{66.8} & \gcell{4065} \\
& \textbf{\textit{w/o}} $\text{MLP}_\text{inter}$ & 
  96.3 & 1334 &
  67.1 & 4007 \\
& \gcell{\textbf{\textit{w/o}} $\text{MLP}_\text{inst}$} & 
  \gcell{96.3} & \gcell{2476} &
  \gcell{67.6} & \gcell{5703} \\
& \ourmethod & 
  96.8 & 1125 &
  67.8 & 3936 \\

\Xhline{1.2pt}
\end{tabular}
}
\vspace{-0.8em}
\caption{Ablation study of \ourmethod.}
\label{tab:ab}
\vspace{-1.8em}
\end{table}

\vspace{-0.5em}
\section{Conclusion}
\vspace{-0.5em}
In this work, we categorize failure modes across different levels and highlight significant gaps in current research. To address these, we conduct a fine-grained neuronal analysis to identify \mon and their fluctuation patterns. Based on these insights, we propose \ourmethod, an explainable, controllable, and unified framework driven by \mon, which demonstrates superior performance and cost-efficiency across multiple domains. We believe it paves the way towards human-like reasoning.

\section*{Limitations}
While \ourmethod demonstrates significant potential in explainability, controllability, and unification, achieving superior performance and token efficiency across diverse tasks and backbone models, we acknowledge certain limitations. The integration of monitoring MLPs incurs inference overhead. Despite implementation optimizations achieving approximate $\mathcal{O}(1)$ time complexity for feature updates ($\blacktriangleright$ Appendix \ref{sec:app-mlp}), marginal latency increases persist; a detailed runtime comparison is provided in Appendix \ref{sec:app-time}. Furthermore, \ourmethod is not currently a fully automated, end-to-end pipeline. We leave the further optimization of this aspect as a direction for future research.

\bibliography{main}

\appendix
\section{Notation}
\label{app:not}
Table \ref{tab:not} summarizes the notations used throughout this paper.

\begin{table*}[t]
\centering
\renewcommand\tabcolsep{10pt} 
\renewcommand\arraystretch{1.2} 

\resizebox{\linewidth}{!}{
\begin{tabular}{c|c}
\Xhline{1.2pt}
\rowcolor{MorandiHeader}
\textbf{Notation} & \textbf{Description} \\
\Xhline{1.2pt}

$L$ & The total number of layers in the Large Language Model (LLM). \\
\gcell{$l^*$} & \gcell{The index of the target intermediate layer for analysis, set to $l^* = L/2$.} \\
$\mathcal{C}$ & The set of all neurons, including both FFN neurons and attention heads, within the target layer $l^*$. \\
\gcell{$y_t$} & \gcell{The target token generated at time step $t$.} \\
$\mathbf{w}_{y_t}$ & The vector representation of the Language Model (LM) head corresponding to token $y_t$. \\
\gcell{$\mathbf{h}_{dec}^{(l^*)}(t, c)$} & \gcell{The decomposed hidden state associated with neuron $c$ at time step $t$ in layer $l^*$.} \\
$\phi(c, t)$ & The attribution score quantifying the independent contribution of neuron $c$ toward the logit of $y_t$. \\
\gcell{$T$} & \gcell{The set of time steps defining the observation window, varying by failure mode level (intra, inter, or instance).} \\
$\mathcal{N}^*$ & The set of significant neurons identified as \mon via the intersection of top-k attribution scores. \\

\gcell{$\mathbf{a}_n$} & \gcell{The sequence of scalar activation values for a neuron $n$ over the time window $T$.} \\
$\mu$ & The mean activation value of the sequence used for zero-centering. \\
\gcell{$P(f)$} & \gcell{The power spectrum derived from the Discrete Fourier Transform (DFT) of the activation sequence.} \\
$\widetilde P(f)$ & The normalized power distribution excluding the Direct Current (DC) component. \\
\gcell{$H$} & \gcell{The normalized spectral entropy quantifying the dispersion of the power spectrum.} \\
$\mathcal{F}_{\mathrm{HF}}$ & The set of high-frequency indices, defined as the latter half of the spectrum. \\
\gcell{$r_{\mathrm{HF}}$} & \gcell{High-frequency energy ratio, a key feature for detecting intra-step failure modes.} \\
$e$ & Total variation energy representing the overall intensity of fluctuations. \\
\gcell{$r_{\mathrm{dom}}$} & \gcell{Dominant frequency energy ratio, indicating the strength of periodic oscillations (inter-step level).} \\
$r_{\mathrm{LF}}$ & Low-frequency energy ratio used to characterize instance-level collapse patterns. \\
\gcell{$\boldsymbol{\phi}_n$} & \gcell{The concatenated input feature vector for neuron $n$ (specific to level $\ell$) used for MLP training.} \\

$\mathcal{D}_{\mathrm{raw}}$ & The raw dataset used for reconstruction, consisting of input-output pairs $(\mathbf{u}, \mathbf{v})$. \\
\gcell{$\widehat{\mathcal{D}}$} & \gcell{The reconstructed dataset containing injected failure modes and trigger mechanisms.} \\
$\mathrm{Seg}(\cdot)$ & The segmentation operator that divides the output $\mathbf{v}$ into step-level subsequences $\pi$. \\
\gcell{$\ell$} & \gcell{The level of failure mode, where $\ell \in \{\mathrm{intra}, \mathrm{inter}\}$.} \\
$\Psi_{\ell}(\cdot)$ & The rewriting operation that injects a specific failure pattern into a critical step. \\
\gcell{$\langle \upsilon_\ell\rangle$} & \gcell{The special trigger token associated with failure level $\ell$, serving as a control signal.} \\
$\mathcal{B}_{\ell}$ & The diagnose-then-correct template containing the prompt $\mathbf{p}_\ell$, diagnosis $\mathbf{d}_\ell$, and correction $\mathbf{c}_\ell$. \\
\gcell{$\theta$} & \gcell{The trainable parameters of the model optimized via SFT.} \\
$\mathcal{I}(\mathbf{u},\widehat{\mathbf{v}})$ & The set of token indices that contribute to the loss function (masking inputs and triggers). \\
\gcell{$x$} & \gcell{The current decoding prefix sequence during inference.} \\
$\tau$ & The length of the current decoding prefix $x$. \\

\Xhline{1.2pt}
\end{tabular}
}
\caption{Comprehensive summary of notations and definitions used in this paper.}
\label{tab:not}
\end{table*}

\section{Algorithm Workflow}
The algorithm framework of \ourmethod is presented in Algorithm \ref{alg}.

\begin{algorithm*}[!t]
\caption{Algorithm workflow of \ourmethod}
\label{alg}
\SetKwInOut{Input}{Input}
\SetKwInOut{Output}{Output}
\SetKw{Break}{break}

\Input{Raw dataset $\mathcal{D}_{\mathrm{raw}}$, LLM parameters $\theta$, Failure Detection MLPs $\mathcal{M}=\{f_{\ell}\}_{\ell}$, Time window $k$}
\Output{Optimized LRM parameters $\theta^*$}

\AlgComment{Phase 1: Pattern Prediction \& Trigger Training} \\
\For{failure level $\ell \in \{\mathrm{intra}, \mathrm{inter}, \mathrm{inst}\}$}{
    Construct feature vectors $\boldsymbol{\phi}^{\ell}$ via Fourier Transform; \SecRef{sec:method-1} \\
    Train lightweight MLP $f_{\ell}$ to predict fluctuation patterns; 
}
Initialize reconstructed dataset $\widehat{\mathcal{D}} \leftarrow \emptyset$\;
\For{$(\mathbf{u}, \mathbf{v}) \in \mathcal{D}_{\mathrm{raw}}$}{
    Segment output $\mathbf{v}$ into steps $\langle \pi_1, \dots, \pi_K \rangle$\;
    Identify critical steps $\mathcal{J}$ for perturbation\;
    \For{$j \in \mathcal{J}$ and level $\ell$}{
        Apply rewriting $\widetilde{\pi}_j^{(\ell)} = \Psi_{\ell}(\mathbf{u}, \pi_{<j}, j)$; \EqRef{eq:rewrite} \\
        Insert trigger $\langle \upsilon_{\ell} \rangle$ and diagnose-then-correct template $\mathcal{B}_{\ell}$; \\
        Construct $\widehat{\mathbf{v}}$ and add $(\mathbf{u}, \widehat{\mathbf{v}})$ to $\widehat{\mathcal{D}}$; \EqRef{eq:reconstruct}
    }
}
\AlgComment{Supervised Fine-Tuning (SFT)} \\
Update $\theta$ by maximizing likelihood on $\widehat{\mathcal{D}}$: \\
$\mathcal{L}_{\mathrm{SFT}}(\theta) = -\mathbb{E}_{(\mathbf{u},\widehat{\mathbf{v}})\sim\widehat{\mathcal{D}}} \sum \log P_\theta(y_t \mid \mathbf{u}, y_{<t})$; \EqRef{eq:sft-block}

\BlankLine
\AlgComment{Phase 2: Inference with Online Monitoring} \\
Given input query $x$, current context $C \leftarrow x$\;
\While{generation not finished}{
    Update sliding window features $\boldsymbol{\phi}$ from recent tokens\;
    \If{$\exists \ell, f_{\ell}(\boldsymbol{\phi})$ detects failure pattern}{
        \AlgComment{Trigger Intervention} \\
        Force next token $y_{next} \leftarrow \langle \upsilon_{\ell} \rangle$\;
        Activate learned behavior $\mathcal{B}_{\ell}$ (Diagnose \& Correct)\;
    }
    \Else{
        Sample next token $y_{next} \sim P_{\theta}(\cdot \mid C)$\;
    }
    Append $y_{next}$ to $C$\;
}
\Return{Final generated reasoning path}
\end{algorithm*}

\section{Related Work}
\label{sec:related}
\paragraph{LLM Reasoning.} According to~\citep{reasoning-survey}, LLM reasoning can be grouped into three families: (1) Deep Reasoning, (2) Feasible Reflection, and (3) Extensive Exploration. \ding{182} \textbf{Deep Reasoning} emphasizes sustaining deep logical processing over long reasoning traces, and is often realized by improving the reasoning format (natural-language, structured/program-like, or latent-space reasoning) and/or by learning long reasoning behaviors from self-generated trajectories. Representative directions include natural-language CoT-style decomposition (e.g., CoT~\citep{cot}, Natural Program~\citep{natural-program}, CodeI/O~\citep{codeio}, CoRT~\citep{cort}), structured and symbolic/programmatic reasoning (e.g., Brain~\citep{brain}, SIaM~\citep{siam}, ENVISIONS~\citep{envisions}, SKIntern~\citep{skintern}, QuaSAR~\citep{quasar}), and latent reasoning mechanisms (e.g., Coconut~\citep{coconut}, Heima~\citep{heima}, LTM~\citep{ltm}), as well as self-learning and tree-search-based training signals (e.g., STaR~\citep{star}, ReST~\citep{rest}, PGTS~\citep{pgts}). \ding{183} \textbf{Feasible Reflection} equips models with iterative self-correction by generating feedback and then performing refinement over earlier reasoning states; typical instantiations range from prompt-based backtracking/refinement to SFT/RL-based~\citep{aurora} reflective learning, with representative methods such as Self-Backtracking~\citep{sb}, Refiner~\citep{refiner}, BackMath~\citep{backmath}, MCTSr~\citep{mctsr}, LLM2~\citep{llm2}, ReARTeR~\citep{rearter} and ReST-MCTS*~\citep{rest-mcts}. \ding{184} \textbf{Extensive Exploration} targets robustness under uncertainty by branching and searching over multiple candidate reasoning trajectories, spanning exploration scaling (sequential vs. parallel), internally learned exploration policies (e.g., STeCa~\citep{steca}), and externally scaffolded search/tool frameworks (e.g. ToT~\citep{tot}, CodeTree~\citep{codetree}, Forest-of-Thought~\citep{fot}). However, most existing works suffer from a lack of explainability, controllability, and unification. In contrast, \ourmethod charts a promising course for future developments in this field.

\paragraph{Interpretability for LLMs.} Interpretability techniques for LRMs span several fronts. \ding{182} \textbf{Attribution-based methods} assign credit to input features or intermediate steps, adapting token-level saliency to trace how each part of a prompt or chain-of-thought contributes to the model’s answer~\citep{attr-1,attr-2,attr-3}. \ding{183} \textbf{Probing techniques} analyze internal representations by training diagnostic classifiers or intervening on hidden states, revealing whether models encode latent reasoning variables~\citep{prob-1,prob-2,prob-3,prob-4}. \ding{184} \textbf{Mechanistic interpretability} directly examines network weights and activations to reverse-engineer the model’s reasoning process, identifying sub-circuits responsible for specific reasoning skills~\citep{mech-1,mech-2,mech-3}. However, there remains a lack of systematic and in-depth mechanistic analysis regarding why LRMs fail during reasoning trajectories. Our work effectively sheds light on these underlying mechanisms.
\begin{algorithm*}[!ht]
\caption{Approximate $O(1)$ Update of Variable-Length Sliding-Window Fourier Features}
\label{alg:o1}
\SetKwInOut{Input}{Input}
\SetKwInOut{Output}{Output}

\Input{Level $\ell\in\{\mathrm{intra},\mathrm{inter}\}$; monitored channels $n\in\{1,\dots,N_{\ell}\}$; new time index $t$ with activations $\{a^{(\ell)}_{n,t}\}$; pop count $r_t\ge0$; fixed probes $\Omega=\{\omega_k\}_{k=1}^{K}$ ($K$ is a small constant); $\varepsilon>0$.}
\Output{Features $\{\boldsymbol{\phi}^{\ell}_n(t)\}_{n=1}^{N_{\ell}}$, where $\boldsymbol{\phi}^{\mathrm{intra}}_n(t)=\big[r_{\mathrm{HF}}^{(n)}(t),H^{(n)}(t),e^{(n)}(t)\big]$ and $\boldsymbol{\phi}^{\mathrm{inter}}_n(t)=\big[r_{\mathrm{dom}}^{(n)}(t),H^{(n)}(t)\big]$.}

\AlgComment{State (persistent across decoding steps). Use $y_{n,i}=|a^{(\ell)}_{n,i}|$.}
Maintain window left index $s$, length $L$; FIFO queue $\mathcal{Q}$ storing vectors $\mathbf{y}_i=[y_{1,i},\dots,y_{N_{\ell},i}]$;
vectors $\mathbf{U}=\sum_{i=s}^{t}\mathbf{y}_i$, $\mathbf{V}=\sum_{i=s}^{t}\mathbf{y}_i\odot\mathbf{y}_i$;
matrix $\mathbf{A}\in\mathbb{C}^{N_{\ell}\times K}$ with $\mathbf{A}_{:,k}=\sum_{i=s}^{t}\mathbf{y}_i\,q_k^{\,i}$.
Maintain phases $\alpha_k=q_k^{t}$, $\beta_k=q_k^{s}$, $\rho_k=q_k^{L}$ and constants $q_k=e^{-i\omega_k}$, $\eta_k=(1-q_k)^{-1}$.

\BlankLine
\AlgComment{1) Push one token (always).}\\
$\mathbf{y}_t \leftarrow [|a^{(\ell)}_{1,t}|,\dots,|a^{(\ell)}_{N_{\ell},t}|]$;\quad Enqueue $\mathbf{y}_t$ into $\mathcal{Q}$;\quad $L\leftarrow L+1$\;
$\mathbf{U}\leftarrow \mathbf{U}+\mathbf{y}_t$;\quad $\mathbf{V}\leftarrow \mathbf{V}+\mathbf{y}_t\odot\mathbf{y}_t$\;
\For{$k\leftarrow 1$ \KwTo $K$}{
    $\alpha_k\leftarrow \alpha_k q_k$;\quad $\rho_k\leftarrow \rho_k q_k$;\quad
    $\mathbf{A}_{:,k}\leftarrow \mathbf{A}_{:,k}+\mathbf{y}_t\,\alpha_k$\;
}

\BlankLine
\AlgComment{2) Pop $r_t$ oldest tokens (optional, can be multiple).}\\
\While{$r_t>0$}{
    Dequeue $\mathbf{y}_s$ from $\mathcal{Q}$;\quad $L\leftarrow L-1$\;
    $\mathbf{U}\leftarrow \mathbf{U}-\mathbf{y}_s$;\quad $\mathbf{V}\leftarrow \mathbf{V}-\mathbf{y}_s\odot\mathbf{y}_s$\;
    \For{$k\leftarrow 1$ \KwTo $K$}{
        $\mathbf{A}_{:,k}\leftarrow \mathbf{A}_{:,k}-\mathbf{y}_s\,\beta_k$;\quad
        $\beta_k\leftarrow \beta_k q_k$;\quad $\rho_k\leftarrow \rho_k q_k^{-1}$\;
    }
    $s\leftarrow s+1$;\quad $r_t\leftarrow r_t-1$\;
}

\BlankLine
\AlgComment{3) Compute features (no traversal over window).}\\
\For{$k\leftarrow 1$ \KwTo $K$}{
    $B_k \leftarrow \beta_k(1-\rho_k)\eta_k$\tcp*{$B_k=\sum_{i=s}^{t}q_k^{\,i}$}
}
\For{$n\leftarrow 1$ \KwTo $N_{\ell}$}{
    $\mu_n \leftarrow U_n/L$\;
    \For{$k\leftarrow 1$ \KwTo $K$}{
        $S_{n,k}\leftarrow A_{n,k}-\mu_n B_k$;\quad $P_{n,k}\leftarrow |S_{n,k}|^2$\;
    }
    $Z_n\leftarrow \sum_{k=1}^{K}P_{n,k}+\varepsilon$;\quad $\widetilde P^{(n)}(k)\leftarrow P_{n,k}/Z_n$\;
    $H^{(n)}(t)\leftarrow -\frac{1}{\log K}\sum_{k=1}^{K}\widetilde P^{(n)}(k)\log\widetilde P^{(n)}(k)$\;

    \uIf{$\ell=\mathrm{intra}$}{
        $r_{\mathrm{HF}}^{(n)}(t)\leftarrow \sum_{k>\lfloor K/2\rfloor}\widetilde P^{(n)}(k)$;\quad
        $E^{(n)}(t)\leftarrow V_n-\frac{U_n^2}{L}$;\quad $e^{(n)}(t)\leftarrow \log(E^{(n)}(t)+\varepsilon)$\;
        $\boldsymbol{\phi}^{\mathrm{intra}}_n(t)\leftarrow \big[r_{\mathrm{HF}}^{(n)}(t),H^{(n)}(t),e^{(n)}(t)\big]$\;
    }
    \Else{
        $r_{\mathrm{dom}}^{(n)}(t)\leftarrow \max_{1\le k\le K}\widetilde P^{(n)}(k)$\;
        $\boldsymbol{\phi}^{\mathrm{inter}}_n(t)\leftarrow \big[r_{\mathrm{dom}}^{(n)}(t),H^{(n)}(t)\big]$\;
    }
}
\Return{$\{\boldsymbol{\phi}^{\ell}_n(t)\}_{n=1}^{N_{\ell}}$}
\end{algorithm*}

\section{Dataset}
\paragraph{MATH.} MATH~\citep{math} is a competition-level mathematics benchmark containing 12,500 problems paired with step-by-step solutions, designed to evaluate multi-step mathematical reasoning beyond routine K–12 exercises. It covers seven subject areas (e.g., algebra, geometry, number theory) and is commonly used for answer exact-match evaluation on the final result.

\paragraph{MATH500.} MATH500~\citep{math500} is a curated subset of 500 representative problems drawn from the MATH benchmark and popularized as a lightweight yet challenging test set for LLM math reasoning. It is widely adopted for standardized comparisons across models.

\paragraph{GSM8K.} GSM8K~\citep{gsm8k} is a dataset of 8.5K human-written grade-school arithmetic word problems, split into 7.5K training and 1K test instances. Problems typically require a short chain of elementary operations (multi-step arithmetic), and performance is usually measured by exact-match accuracy on the final answer.

\paragraph{AIME24.} AIME24~\citep{aime} is an olympiad-style math benchmark constructed from the 2024 American Invitational Mathematics Examination (AIME) I and II, totaling 30 problems. Each question expects an integer answer in [0,999], enabling reliable automatic evaluation via exact match while still demanding substantial symbolic and combinational reasoning.

\paragraph{AIME25.} AIME25~\citep{aime} similarly aggregates 30 problems from the 2025 AIME I and II into an evaluation set with integer answers in [0,999]. Owing to its ``fresh'' annual release and contest difficulty, AIME25 is frequently used as a stringent test of advanced mathematical reasoning and generalization under minimal ambiguity in answer format.

\paragraph{GPQA-Diamond.} GPQA~\citep{gpqa} is a graduate-level, ``Google-proof'' multiple-choice QA benchmark authored and validated by domain experts across biology, chemistry, and physics; the Diamond split is a higher-quality, more challenging subset containing 198 questions. It is commonly used to probe scientific reasoning under expert-level knowledge demands, with accuracy computed over 4-option multiple-choice answers.

\paragraph{LiveCodeBench.} LiveCodeBench~\citep{livecodebench} is a continuously updated coding benchmark explicitly designed to mitigate test-set contamination by collecting newly released competitive-programming problems over time. Beyond code generation, it emphasizes holistic coding abilities (e.g., self-repair, execution, test-output prediction) and provides time-stamped releases (e.g., hundreds of problems spanning May 2023 onward) for reproducible evaluation.

\paragraph{Bespoke-Stratos-17k.} Bespoke-Stratos-17k~\citep{bs17k} is a reasoning distillation dataset (17K examples) consisting of questions paired with reasoning traces and final answers, created by replicating and improving the Berkeley Sky-T1 pipeline using distillation data from DeepSeek-R1. It is used as supervised fine-tuning data to induce long-form, explicit reasoning behaviors across domains including math and coding.

\section{Technical Details}
\label{sec:app-tech}
\subsection{Data Preprocessing}
Following LightThinker~\citep{lightthinker}, we utilize BS-17k~\citep{bs17k} as the training dataset for both MLPs and SFT. To prevent potential data leakage, we explicitly exclude samples overlapping with our evaluation benchmarks (MATH500, GSM8K, AIME24, AIME25, GPQA-Diamond, and LiveCodebench) via text matching.

\begin{table}[t]
\centering

\resizebox{0.9\linewidth}{!}{
    \begin{tabular}{c|cc}
    \Xhline{1.2pt}
    \rowcolor{MorandiHeader}
    \textbf{Failure Mode} & \textbf{Accuracy} & \textbf{Recall} \\
    \Xhline{1.2pt}
    
    $\text{MLP}_{\mathrm{intra}}$ & $0.871\pm0.010$ & $0.848 \pm 0.012$ \\
    
    \gcell{$\text{MLP}_{\mathrm{inter}}$} & \gcell{$0.862\pm0.011$} & \gcell{$0.936\pm0.008$} \\
    
    $\text{MLP}_{\mathrm{inst}}$ & $0.944\pm0.006$ & $0.951\pm0.007$ \\
    \Xhline{1.2pt}
    \end{tabular}
}
\caption{Experimental results across different MLPs.}
\label{tab:mlp}
\end{table}

\subsection{Elaboration on \mon}
As discussed in the Limitations, \ourmethod is not currently a fully automated, end-to-end framework, primarily due to variations in layer depth and architecture across different backbones. However, we emphasize that while the specific constituent neurons of \mon vary, they adhere to universal patterns established in Section \ref{sec:dive}. Specifically: \ding{182} intra-step and instance-level \mon are predominantly located in FFNs, whereas inter-step \mon reside in attention heads; and \ding{183} intra-step failures exhibit significant fluctuation amplitudes, inter-step failures display periodicity, and instance-level failures are characterized by sudden collapse. In practice, this necessitates independent \mon extraction and MLP training for each backbone. Fortunately, these processes remain lightweight. For instance, training the MLPs requires labeling only $\sim20$ reasoning trajectories, which yields approximately $\sim5,000$ token-level training samples. Extending this framework to achieve cross-backbone alignment represents a pivotal direction for our future work.

\begin{table*}[t]
\centering
\renewcommand\tabcolsep{4pt}
\renewcommand\arraystretch{1.1}

\resizebox{\linewidth}{!}{
\begin{tabular}{l|cc|cc|cc|cc|cc|cc}
\Xhline{1.2pt}
\rowcolor{MorandiHeader}
& \multicolumn{2}{c|}{\textbf{R1-Qwen-7B}} & \multicolumn{2}{c|}{\textbf{R1-Qwen-32B}} 
& \multicolumn{2}{c|}{\textbf{Qwen3-8B-thinking}} & \multicolumn{2}{c|}{\textbf{Qwen3-32B-thinking}} 
& \multicolumn{2}{c|}{\textbf{R1-Llama-8B}} & \multicolumn{2}{c}{\textbf{R1-Llama-70B}} \\
\rowcolor{MorandiHeader}
\multirow{-2}{*}{\textbf{Method}} 
& \textbf{Pass@1} & \textbf{Token} & \textbf{Pass@1} & \textbf{Token} 
& \textbf{Pass@1} & \textbf{Token} & \textbf{Pass@1} & \textbf{Token} 
& \textbf{Pass@1} & \textbf{Token} & \textbf{Pass@1} & \textbf{Token} \\
\Xhline{1.2pt}

Vanilla 
& 54.4 & 10438 & 70.0 & 7873 
& 70.0 & 11125 & \underline{77.8} & 10677 
& 45.6 & 10798.9 & 68.9 & 7766 \\

\gcell{DAST} 
& \gcell{55.6} & \gcell{7258} & \gcell{70.0} & \gcell{5802} 
& \gcell{67.8} & \gcell{\underline{5964}} & \gcell{\underline{77.8}} & \gcell{\textbf{5981}} 
& \gcell{45.6} & \gcell{8246} & \gcell{67.8} & \gcell{5115} \\

Think or Not 
& 52.2 & \textbf{4341} & 67.8 & \textbf{3993} 
& 68.9 & \textbf{5387} & 76.7 & 6173 
& 44.4 & 6761 & 71.1 & \textbf{4005} \\

\gcell{AlphaOne} 
& \gcell{55.6} & \gcell{8224} & \gcell{72.2} & \gcell{8210} 
& \gcell{\underline{73.3}} & \gcell{8343} & \gcell{\textbf{78.9}} & \gcell{8007} 
& \gcell{\underline{47.8}} & \gcell{8339} & \gcell{72.2} & \gcell{7873} \\

RL + LP 
& 52.2 & 5693 & 71.1 & 5492 
& 71.1 & 6986 & 76.7 & 6238 
& 45.6 & 5333 & 66.7 & 5304 \\

\gcell{GRPO} 
& \gcell{\underline{56.7}} & \gcell{11673} & \gcell{\underline{73.3}} & \gcell{8389} 
& \gcell{72.2} & \gcell{10931} & \gcell{\textbf{78.9}} & \gcell{11934} 
& \gcell{44.4} & \gcell{11312} & \gcell{72.2} & \gcell{8109} \\

S-GRPO 
& 54.4 & 5094 & 70.0 & 4906 
& \underline{73.3} & 6771 & \underline{77.8} & 6040 
& 45.6 & \textbf{4809} & 70.0 & \underline{5002} \\

\gcell{DAPO} 
& \gcell{\textbf{57.8}} & \gcell{11908} & \gcell{\underline{73.3}} & \gcell{8817} 
& \gcell{71.1} & \gcell{11781} & \gcell{\textbf{78.9}} & \gcell{12038} 
& \gcell{46.7} & \gcell{12079} & \gcell{\underline{73.3}} & \gcell{8589} \\

\ourmethod 
& \textbf{57.8} & \underline{4997} & \textbf{74.4} & \underline{4456} 
& \textbf{74.4} & 6609 & \textbf{78.9} & \underline{6029} 
& \textbf{50.0} & \underline{5116} & \textbf{74.4} & 5237 \\

\dcell{$\Delta$} 
& \dcell{\darkred{\textbf{+3.4}}} & \dcell{\darkblue{\textbf{52.1\%}}} & \dcell{\darkred{\textbf{+4.4}}} & \dcell{\darkblue{\textbf{43.4\%}}} 
& \dcell{\darkred{\textbf{+4.4}}} & \dcell{\darkblue{\textbf{40.6\%}}} & \dcell{\darkred{\textbf{+1.1}}} & \dcell{\darkblue{\textbf{43.5\%}}} 
& \dcell{\darkred{\textbf{+4.4}}} & \dcell{\darkblue{\textbf{52.6\%}}} & \dcell{\darkred{\textbf{+5.5}}} & \dcell{\darkblue{\textbf{32.6\%}}} \\

\Xhline{1.2pt}
\end{tabular}
}
\caption{Additional results on the AIME24 dataset. Best results are \textbf{bold}, runners-up \underline{underlined}.}
\label{tab:aime24}
\end{table*}

\begin{table*}[t]
\centering
\renewcommand\tabcolsep{6pt}
\renewcommand\arraystretch{1.2}

\resizebox{\linewidth}{!}{
\begin{tabular}{c|l|cccccc}
\Xhline{1.2pt}
\rowcolor{MorandiHeader}
\textbf{Model} & \textbf{Method} &
\textbf{AIME24} & \textbf{AIME25} & \textbf{MATH500} & \textbf{GSM8K} & \textbf{GPQA-Diamond} & \textbf{LiveCodeBench} \\
\Xhline{1.2pt}

\multirow{2}{*}{DeepSeek-R1-Qwen-7B}
& Vanilla &
  02:13 & 02:12 & 04:11 & 02:26 & 04:52 & 24:20 \\
& \gcell{\ourmethod} &
  \gcell{02:24} & \gcell{02:23} & \gcell{04:29} & \gcell{02:38} & \gcell{05:13} & \gcell{25:51} \\
\hline

\multirow{2}{*}{DeepSeek-R1-Qwen-32B}
& Vanilla &
  04:00 & 04:10 & 07:06 & 01:45 & 07:29 & 36:36 \\
& \gcell{\ourmethod} &
  \gcell{04:17} & \gcell{04:28} & \gcell{07:35} & \gcell{01:54} & \gcell{07:59} & \gcell{38:51} \\
\hline

\multirow{2}{*}{Qwen3-8B-thinking}
& Vanilla &
  02:26 & 02:28 & 06:24 & 05:47 & 04:12 & 24:56 \\
& \gcell{\ourmethod} &
  \gcell{02:38} & \gcell{02:40} & \gcell{06:50} & \gcell{06:11} & \gcell{04:30} & \gcell{26:29} \\
\hline

\multirow{2}{*}{Qwen3-32B-thinking}
& Vanilla &
  04:25 & 04:34 & 10:04 & 08:09 & 06:21 & 44:15 \\
& \gcell{\ourmethod} &
  \gcell{04:44} & \gcell{04:53} & \gcell{10:43} & \gcell{08:41} & \gcell{06:47} & \gcell{46:57} \\
\hline

\multirow{2}{*}{DeepSeek-R1-Llama-8B}
& Vanilla &
  02:10 & 02:06 & 04:49 & 02:38 & 04:59 & 25:21 \\
& \gcell{\ourmethod} &
  \gcell{02:21} & \gcell{02:17} & \gcell{05:09} & \gcell{02:50} & \gcell{05:20} & \gcell{26:55} \\
\hline

\multirow{2}{*}{DeepSeek-R1-Llama-70B}
& Vanilla &
  06:41 & 06:37 & 10:47 & 03:42 & 10:25 & 57:56 \\
& \gcell{\ourmethod} &
  \gcell{07:08} & \gcell{07:04} & \gcell{11:29} & \gcell{03:58} & \gcell{11:06} & \gcell{61:28} \\

\Xhline{1.2pt}
\end{tabular}
}
\caption{Inference efficiency comparison (Time in MM:SS).}
\label{tab:app-time}
\end{table*}

\begin{table}[!t]
\centering
\resizebox{0.6\linewidth}{!}{
\begin{tabular}{l|c|cc}
\Xhline{1.2pt}
\rowcolor{MorandiHeader}
\textbf{Hyper} & \textbf{Value} & \textbf{Pass@1} & \textbf{Token} \\
\Xhline{1.2pt}

\multirow{3}{*}{$k_{\text{intra}}$}
& 2 & 96.0 & 1360 \\
& \gcell{4} & \gcell{96.3} & \gcell{1318} \\
& 8 & 96.2 & 1335 \\
\hline

\multirow{3}{*}{$k_{\text{inter}}$}
& 2 & 96.1 & 1345 \\
& \gcell{4} & \gcell{96.3} & \gcell{1318} \\
& 8 & 96.4 & 1328 \\
\hline

\multirow{3}{*}{$K_{\text{inst}}$}
& 2 & 95.9 & 1180 \\
& \gcell{4} & \gcell{96.3} & \gcell{1318} \\
& 8 & 96.5 & 1575 \\

\Xhline{1.2pt}
\end{tabular}
}
\caption{Hyperparameter sensitivity on MATH500 with DeepSeek-R1-Distill-Qwen-32B.}
\label{tab:sens}
\end{table}

\subsection{MLPs Training}
\label{sec:app-mlp}
\paragraph{Datasets.} We utilize the BS-17k dataset for MLP training, randomly partitioned into training and test sets at an $8:2$ ratio. To maximize the diversity of erroneous reasoning paths, we perform repeated sampling using DeepSeek-R1-Distill-Qwen-7B and DeepSeek-R1-Distill-Llama-8B to collect reasoning trajectories with associated neuron activations. We then filter for failure instances and employ an LLM-as-a-Judge approach to annotate the specific token segments corresponding to these failures.

\paragraph{Training Details.} We adopt a unified architecture for MLPs across all levels: a simple three-layer fully connected network, with input dimensions determined by the respective input features. Training employs the GELU activation function, Dropout regularization, binary cross-entropy loss, and the AdamW optimizer.

\paragraph{Results.} Table \ref{tab:mlp} presents the experimental results on the test set, reporting accuracy and recall. The results demonstrate the excellent performance of our trained MLPs.

\paragraph{Feature Update.} Algorithm \ref{alg:o1} presents our optimized sliding-window feature update procedure, achieving $\mathcal{O}(1)$ time complexity.

\subsection{Fine-tuning}
\label{sec:app-ft}
We employ full-parameter fine-tuning for $1$ epoch, utilizing a cosine warmup schedule with a ratio of $0.05$. The learning rate is set to $1\text{e-}5$, and the batch size is $64$.

\subsection{Baseline Settings}
For all baselines, we strictly adhere to the settings detailed in their respective original papers. To ensure fair comparison, we standardize generation hyperparameters across all methods: temperature is set to $0.6$, top-$p$ to $0.95$, and the maximum token limit to $16,000$. We report the average Pass@1 accuracy over five independent runs. Additionally, for RL-based baselines, experiments are conducted on A100 GPUs utilizing bf16 mixed precision and DeepSpeed ZeRO-3.

\section{Additional Results}
\label{sec:add-re}
\subsection{AIME24 Results}
Table \ref{tab:aime24} presents additional experimental results on the AIME24 dataset.

\subsection{Runtime Comparison}
\label{sec:app-time}
Table \ref{tab:app-time} presents the inference runtime comparison between \ourmethod and the Vanilla model, measured on A100 GPU. It is evident that while \ourmethod introduces marginal overhead, it remains within an acceptable range.

\subsection{Hyperparameter Sensitivity}
\label{sec:app-sens}
We present the results of the hyperparameter sensitivity experiments in Table \ref{tab:sens}.

\section{Prompt \& Example Data}
\label{sec:prompt}
We provide all prompts utilized in this paper, including those for dataset reconstruction and failure mode detection, alongside examples of reconstructed data.


\begin{PromptFrame}{\textbf{\textsc{Intra-step Level Reconstruction}}}
\begin{lstlisting}[style=PromptStyle]
### SYSTEM
You are a data-construction assistant for SFT. Your task is to generate reasoning trajectories containing a specific "Intra-step Level" failure and diagnose-then-correct pattern.

### INPUT FORMAT
1. <PROBLEM> ... </PROBLEM>
2. <REFERENCE_REASONING> ... </REFERENCE_REASONING>
3. <CONTROL>
   error_length: {short|medium|long}
   error_type: {dropped_case|invalid_division|sign_error|algebra_simplification_error|mistaken_assumption|domain_violation}
   style: LRM_natural_first_person
   </CONTROL>

### INSTRUCTIONS
Produce a single reasoning trace following these steps strictly:
1. **Context:** Copy the first 2-4 steps of the <REFERENCE_REASONING> verbatim.
2. **Error Injection:** Identify the next critical step and rewrite it to be plausibly *wrong* based on the `error_type`. Do not introduce multiple errors.
3. **Trigger Insertion:** Immediately after the wrong step, insert this exact block:
   <INTRA>
   [FAILURE_INTRA] A key reasoning error surfaced a few steps back. I should identify the mistaken span, briefly analyze it, and correct it before continuing.
   - **error span**: "<quote the wrong text span>"
   - **analysis**: <1-3 sentences explaining why it is wrong>
4. **Correct & Continue:** After the block, "back up" logically, apply the correction, and continue reasoning to the correct final answer.
5. **Format:** Use natural first-person reasoning. Return ONLY the reconstructed reasoning trace (no meta commentary, no JSON, no extra headers).
\end{lstlisting}
\end{PromptFrame}

\begin{PromptFrame}{\textbf{\textsc{Inter-step Level Reconstruction}}}
\begin{lstlisting}[style=PromptStyle]
### SYSTEM
You are a data-construction assistant for SFT. Your task is to generate reasoning trajectories containing an "Inter-step Level" stagnation-and-repair pattern.

### INPUT FORMAT
1. <PROBLEM> ... </PROBLEM>
2. <REFERENCE_REASONING> ... </REFERENCE_REASONING>
3. <CONTROL>
   loop_length_paragraphs: {3|4|5}
   loop_theme: {mod_checks|bounds|symmetry_observations|equivalent_reformulations}
   style: LRM_natural_first_person
   </CONTROL>

### INSTRUCTIONS
Produce a single reasoning trace following these steps strictly:
1. **Context:** Copy the first 2-4 steps of the <REFERENCE_REASONING> verbatim.
2. **Loop Generation:** Create a realistic stagnation loop of `loop_length_paragraphs`.
   - The model must oscillate between strategies related to `loop_theme`.
   - It should sound logical but fail to make decisive progress (spinning wheels).
   - Do not generate nonsense; simulate a model trying but failing to break through.
3. **Trigger Insertion:** Immediately after the loop, insert this exact block:
   <INTER>
   [FAILURE_INTER] I'm stuck in an inter-step loop: I keep revisiting the same near equivalent checks and switching between them whenever one stalls, without any progress. Let me summarize the loop, then pivot to a genuinely different route.
   - **past attempts summary:** <2-4 bullet-worthy clauses summarizing the repeated approaches>
   - **problem-grounded analysis:** <1-3 sentences explaining why those moves aren't breaking the core constraint>
4. **Pivot & Finish:** Define a pivot plan (1-2 sentences naming a different framework and the next concrete step). Insert this plan into the block above, then immediately execute it and continue reasoning to the correct final answer.
5. **Format:** Use natural first-person reasoning. Return ONLY the reconstructed reasoning trace (no meta commentary, no JSON, no extra headers).
\end{lstlisting}
\end{PromptFrame}

\begin{PromptFrame}{\textsc{Intra-step Level Failure Detection}}
\begin{lstlisting}[style=PromptStyle]
### SYSTEM
You are an expert Logic Verifier. Your goal is to strictly evaluate a reasoning trace for **Local Validity**.

You will be given a <PROBLEM> and a <REASONING_TRACE>. You must verify the trace step-by-step.

### DETECTION CRITERIA
Flag a step as an [ERROR] if it contains any of the following "Intra-step" failures:
1. **Calculation/Algebra**: Sign errors, invalid simplification, arithmetic mistakes (e.g., 2+2=5).
2. **Domain Violation**: Dividing by zero, taking the log of a negative, applying a theorem outside its valid conditions.
3. **Logic Non-sequitur**: The conclusion of step N does not logically follow from step N-1.
4. **Hallucination**: Inventing constraints or values not present in the problem context.
5. **Dropped Case**: Arbitrarily narrowing the scope (e.g., assuming x is positive without proof).

**CRITICAL NOTE**: 
- Ignore "inefficient" steps or "circular" reasoning (that is a separate check). 
- Focus ONLY on whether the specific statement is *factually* or *mathematically* false.

### OUTPUT FORMAT
Return valid JSON only.

{
  "has_error": boolean,
  "first_error_step": string | null, // Quote the specific sentence/equation
  "error_type": "calculation" | "logic_non_sequitur" | "domain_violation" | "hallucination" | "dropped_case" | "none",
  "explanation": "Brief analysis of why this step is invalid given the prior context."
}

### INPUT
<PROBLEM>
{{PROBLEM_TEXT}}
</PROBLEM>

<REASONING_TRACE>
{{MODEL_OUTPUT}}
</REASONING_TRACE>
\end{lstlisting}
\end{PromptFrame}

\begin{PromptFrame}{\textsc{Inter-step Level Failure Detection}}
\begin{lstlisting}[style=PromptStyle]
### SYSTEM
You are an expert Metacognitive Strategy Evaluator. Your goal is to analyze the **Global Information Flow** of a reasoning trace to detect Stagnation or Loops. You will be given a <PROBLEM> and a <REASONING_TRACE>. Do not check for minor calculation errors. Instead, check if the reasoning is "spinning its wheels."

### DETECTION CRITERIA
Flag the trace as [STAGNANT] if it exhibits these "Inter-step" patterns for 3+ consecutive paragraphs/steps:
1. **Equivalent Reformulations**: Rewriting the same equation in different forms without isolating new variables (e.g., x=5-y -> y=5-x -> x+y=5).
2. **Strategy Oscillation**: Switching back and forth between two approaches (e.g., Modular Arithmetic -> Bounds -> Modular Arithmetic) without ruling anything out.
3. **Empty Verbosification**: Long explanations that restate the goal or definitions without deriving new data.
4. **Repetitive Checks**: Testing values or cases that were already implicitly or explicitly handled.

**CRITICAL NOTE**:
- A long derivation is NOT a loop if it is making progress towards a solution.
- A "stagnation" means the *Information Entropy* is not decreasing (the search space isn't shrinking).

### OUTPUT FORMAT
Return valid JSON only.

{
  "is_stagnant": boolean,
  "loop_detected_at": "Approximate start of the loop (e.g., 'Paragraph 4')", 
  "loop_theme": "modular_checks" | "bounds_restatement" | "equation_shuffling" | "none",
  "severity": "low" | "medium" | "high" | "none",
  "explanation": "Brief analysis of the repetitive pattern observed."
}

### INPUT
<PROBLEM>
{{PROBLEM_TEXT}}
</PROBLEM>

<REASONING_TRACE>
{{MODEL_OUTPUT}}
</REASONING_TRACE>
\end{lstlisting}
\end{PromptFrame}

\begin{PromptFrame}{\textsc{Example of Intra-step Level Data}}
\begin{lstlisting}[style=PromptStyle]
[Question]:
How many positive integers n satisfy that n^2 + 2n + 2017 is a perfect square?

[Response]:
Let n^2 + 2n + 2017 = m^2 for some integer m.

Complete the square on n: n^2 + 2n + 2017 = (n+1)^2 + 2016, so we need
m^2 = (n+1)^2 + 2016.

Rearrange as a difference of squares:
m^2 - (n+1)^2 = 2016
(m-(n+1))(m+(n+1)) = 2016.

Let u = m-(n+1) and v = m+(n+1). Then u and v are positive integers with uv = 2016 and v > u, and we can recover
n+1 = (v-u)/2.

At first glance it seems each factor pair (u,v) with uv=2016 and u<v gives a valid n, so the number of solutions should just be the number of positive factor pairs of 2016, which is d(2016)/2 = 36/2 = 18.

<INTRA>
[FAILURE_INTRA] A key reasoning error occurred in the counting step; I need to pinpoint it and fix it before continuing.
- **error span**: "each factor pair (u,v) with uv=2016 and u<v gives a valid n ... so ... d(2016)/2 = 18."
- **analysis**: The recovery formula n+1 = (v-u)/2 requires (v-u) to be even, i.e., u and v must have the same parity. This was not checked. Since uv=2016 is even, u and v cannot both be odd; the only way to have the same parity is that both are even. Counting all factor pairs overcounts pairs with opposite parity, which do not yield an integer n.

Back up and apply the parity constraint properly. Because n+1 = (v-u)/2 must be an integer, u and v must have the same parity. With uv=2016 even, "same parity" forces u and v to both be even.

So write u = 2u' and v = 2v'. Then uv = 2016 becomes 4u'v' = 2016, hence
u'v' = 504,
and still v>u corresponds to v' > u'.

Now each factor pair (u',v') with u'v'=504 and u'<v' gives
n+1 = (v-u)/2 = (2v' - 2u')/2 = v' - u',
so n = (v' - u') - 1.

We need n > 0, i.e. v' - u' > 1. The excluded case v' - u' = 1 would require u'(u'+1)=504, but 504 is not a product of consecutive integers (the discriminant 1+4*504=2017 is not a perfect square), so every valid factor pair with u'<v' gives n>0.

Therefore the number of valid n equals the number of factor pairs u'<v' of 504.

Factor 504 = 2^3 * 3^2 * 7, so the number of divisors is
(3+1)(2+1)(1+1) = 4*3*2 = 24.
Since 504 is not a perfect square, the number of unordered factor pairs is 24/2 = 12, and that equals the number of pairs with u'<v'.

Hence there are 12 positive integers n satisfying the condition.

Final Answer: 12.
\end{lstlisting}
\end{PromptFrame}

\begin{PromptFrame}{\textsc{Example of Inter-step Level Data}}
\begin{lstlisting}[style=PromptStyle]
[Question]:
How many ordered pairs of integers (x, y) satisfy x^2 + xy + y^2 = 91?

[Response]:
This is a positive definite quadratic form, so there will only be finitely many integer solutions. I'll start by bounding the size.

Since x^2 + xy + y^2 >= (1/2)(x^2 + y^2) (because the smallest eigenvalue of the symmetric form is 1/2), we get
(1/2)(x^2 + y^2) <= 91
so x^2 + y^2 <= 182, hence |x|,|y| <= 13.

A direct search in that box is feasible in principle, but I'd like a cleaner counting method.

Treat the equation as a quadratic in x:
x^2 + yx + (y^2 - 91) = 0.
For x to be an integer, the discriminant must be a perfect square:
Delta = y^2 - 4(y^2 - 91) = 364 - 3y^2.

So I need 364 - 3y^2 to be a nonnegative square. That already bounds y:
3y^2 <= 364 => |y| <= 11.

At this point I could try listing y = -11,...,11 and checking which values make Delta a square, but I'm also tempted to see if modular arithmetic can prune first.

Modulo 3, x^2 + xy + y^2 == 91 == 1 (mod 3). The expression can be 0 or 1 mod 3 depending on (x,y), so that's not restrictive. Mod 4 doesn't help much either since squares are 0 or 1 and xy mixes parities.

Maybe I should exploit the discriminant form: Delta = 364 - 3y^2. If I set Delta = k^2, then 3y^2 + k^2 = 364. That resembles a sum of squares condition, but I'm not sure how to count it cleanly without still enumerating y.

I can also swap roles: solving for y gives the same discriminant condition, so it doesn't change the nature of the search.

Let me try more modular filtering: since Delta is a square, Delta == 0 or 1 (mod 4). But 364 - 3y^2 == 0 - 3y^2 (mod 4) == -3y^2 (mod 4). If y is even, y^2 == 0 so Delta == 0 (ok). If y is odd, y^2 == 1 so Delta == -3 == 1 (ok). So parity doesn't prune.

Trying mod 8: if y is odd, 3y^2 == 3 (mod 8), so Delta == 364 - 3 == 1 (mod 8), which is a square residue. If y even, Delta == 4 (mod 8), also a square residue. So again permissive.

I'm going in circles: each "check" restates the same discriminant condition or gives permissive residue constraints. The cleanest might just be to enumerate y and count x from the quadratic formula, but I'd like a method that feels more structural.

<INTER>
[FAILURE_INTER] I'm stuck in an inter-step loop: I keep revisiting the same near equivalent checks and switching between them whenever one stalls, without any progress. Let me summarize the loop, then pivot to a genuinely different route.
- **past attempts summary:** I bounded |x|,|y| and rewrote as a quadratic in x with discriminant Delta=364-3y^2; then I repeatedly tried modular residue checks (mod 3/4/8) and symmetry swaps, but each move either restated the same discriminant constraint or stayed too permissive to reduce the search.
- **problem-grounded analysis:** The form x^2+xy+y^2 has a standard linear change of variables that diagonalizes it: 4(x^2+xy+y^2) = (2x+y)^2 + 3y^2. Converting the problem to u^2 + 3v^2 = 364 (with a simple parity condition) turns it into a clean "square checking" enumeration over a much smaller set with no looping.

I'll take that diagonalization route and keep it consistent.

Start from x^2 + xy + y^2 = 91 and multiply by 4:
4x^2 + 4xy + 4y^2 = 364.

Notice (2x+y)^2 = 4x^2 + 4xy + y^2, so
4x^2 + 4xy + 4y^2 = (2x+y)^2 + 3y^2.

Let u = 2x + y and v = y. Then the equation becomes
u^2 + 3v^2 = 364.

Also, since x = (u - v)/2, we need u - v to be even (equivalently u == v mod 2) so that x is an integer.

Now bound v:
3v^2 <= 364 => |v| <= 11.

I can now check v = -11,...,11 and see when 364 - 3v^2 is a perfect square u^2, while also ensuring u == v (mod 2).

Compute:
- v = +/-11: 364 - 3*121 = 364 - 363 = 1 => u = +/-1 (odd; matches v odd).
- v = +/-10: 364 - 3*100 = 64 => u = +/-8 (even; matches v even).
- v = +/-9:  364 - 3*81  = 121 => u = +/-11 (odd; matches v odd).
- v = +/-6:  364 - 3*36  = 256 => u = +/-16 (even; matches v even).
- v = +/-5:  364 - 3*25  = 289 => u = +/-17 (odd; matches v odd).
- v = +/-1:  364 - 3*1   = 361 => u = +/-19 (odd; matches v odd).

All other v in -11...11 give non-squares, and in every listed case the parity condition u == v (mod 2) is satisfied automatically.

For each absolute value |v| in {1,5,6,9,10,11}, there are:
- 2 choices for sign of v,
- 2 choices for sign of u,
giving 4 solutions (u,v).

That's 6 * 4 = 24 solutions (u,v), and each corresponds to a unique integer pair (x,y) via
y = v, x = (u - v)/2.

Therefore, the number of ordered integer pairs (x,y) is 24.

Final Answer: 24
\end{lstlisting}
\end{PromptFrame}


\end{document}